\RequirePackage{fix-cm}
\documentclass[twocolumn]{svjour3}          % twocolumn
\smartqed  % flush right qed marks, e.g. at end of proof
\usepackage{amsmath,amsfonts,bm}

\def\eqref#1{equation~\ref{#1}}
\def\1{\bm{1}}

\DeclareMathAlphabet{\mathsfit}{\encodingdefault}{\sfdefault}{m}{sl}
\SetMathAlphabet{\mathsfit}{bold}{\encodingdefault}{\sfdefault}{bx}{n}
\usepackage{hyperref}
\usepackage{url}

\usepackage{booktabs}                  % professional-quality tables
\usepackage{amsfonts,amsmath,amssymb}  % blackboard math symbols
\usepackage{nicefrac}                  % compact symbols for 1/2, etc.
\usepackage{microtype}                 % microtypography
\usepackage{color}
\usepackage{graphicx,float,subfigure,caption}
\newcommand{\z}{{\rm\bf z}}       % notation of latent code.
\newcommand{\Z}{\mathcal{Z}}      % notation of latent space.
\newcommand{\w}{{\rm\bf w}}       % notation of disentangled latent code.
\newcommand{\W}{\mathcal{W}}      % notation of disentangled latent space.
\newcommand{\y}{{\rm\bf y}}       % notation of style concept.
\newcommand{\A}{{\rm\bf A}}       % notation of weight of style transformation.
\renewcommand{\b}{{\rm\bf b}}     % notation of bias of style transformation.
\newcommand{\x}{{\rm\bf x}}       % notation of image.
\newcommand{\X}{\mathcal{X}}      % notation of image space.
\renewcommand{\S}{\mathcal{S}}    % notation of semantic space.
\newcommand{\n}{{\rm\bf n}}       % notation of normal direction of boundary.

\usepackage{times}
\usepackage{epsfig}
\usepackage{graphicx}
\usepackage[fleqn]{amsmath}
\usepackage{amssymb,amsfonts,dsfont,bm,mathrsfs}
\usepackage{booktabs,multirow,adjustbox}
\usepackage{url}
\usepackage[numbers]{natbib}
\usepackage{color}
\usepackage{float}
\begin{document}

\title{Semantic Hierarchy Emerges in Deep Generative Representations for Scene Synthesis}



\author{
  Ceyuan Yang* \thanks{* denotes equal contribution.}  \and
  Yujun Shen*   \and
  Bolei Zhou
}

\authorrunning{Yang et al.} % if too long for running head

\institute{
  C. Yang, Y. Shen, B. Zhou \at
  \small{\texttt{\{yc019,sy116,bzhou\}@ie.cuhk.edu.hk}} \\
  Department of Information Engineering, The Chinese University of Hong Kong, Hong Kong.
}


\maketitle

\begin{abstract}
  % !TEX root = main.tex

Despite the success of Generative Adversarial Networks (GANs) in image synthesis, there lacks enough understanding on what generative models have learned inside the deep generative representations and how photo-realistic images are able to be composed of the layer-wise stochasticity introduced in recent GANs.
In this work, we show that highly-structured semantic hierarchy emerges as variation factors from synthesizing scenes from the generative representations in state-of-the-art GAN models, like StyleGAN and BigGAN.
By probing the layer-wise representations with a broad set of semantics at different abstraction levels, we are able to \emph{quantify} the causality between the activations and semantics occurring in the output image.
Such a quantification identifies the human-understandable variation factors learned by GANs to compose scenes.
The qualitative and quantitative results further suggest that the generative representations learned by the GANs with layer-wise latent codes are specialized to synthesize different hierarchical semantics: the early layers tend to determine the spatial layout and configuration, the middle layers control the categorical objects, and the later layers finally render the scene attributes as well as color scheme.
Identifying such a set of manipulatable latent variation factors facilitates semantic scene manipulation.\footnote{Code and demo video can be found at \url{https://ceyuan.me/SemanticHierarchyEmerge}.}

  \vspace{5ex}
\end{abstract}

% !TEX root = main.tex

\begin{figure*}[t]
  \centering
  \includegraphics[width=1.0\textwidth]{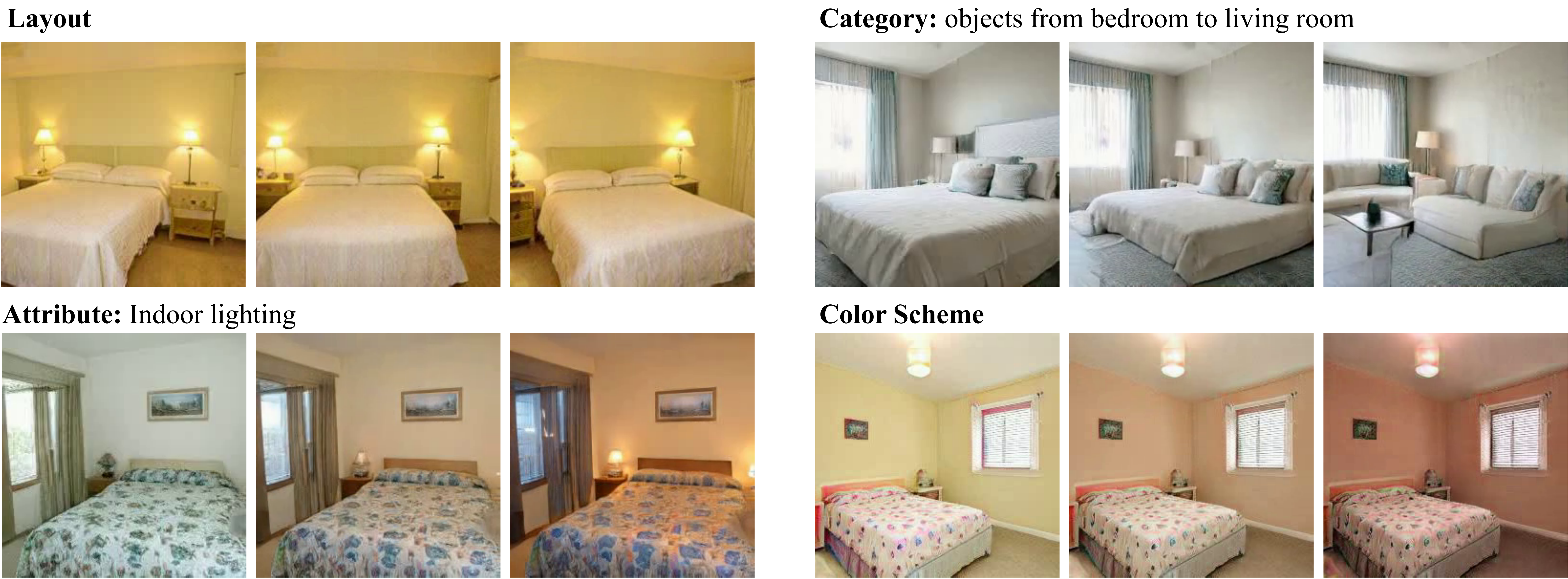}
  \captionsetup{font=small}
  \caption{
    Manipulation results from four different abstraction levels, including \emph{layout}, \emph{categorical objects}, \emph{scene attributes}, and \emph{color scheme}. For each tuple of images, the first is the original synthesized image, the following are the ones after some degree of manipulation.
  }
  \label{fig:teaser}
\end{figure*}

\section{Introduction}

Success of deep neural networks stems from the representation learning, which identifies the explanatory factors underlying the high-dimensional observed data \cite{bengio2013representation}.
Prior work has shown that many concept detectors spontaneously emerge inside the deep representations trained for the classification task.
For example, \citet{gonzalez2018semantic} shows that networks for object recognition are able to detect semantic object parts, and \citet{bau2017network} confirms that deep representations from classifying images learn to detect different categorical concepts at different layers.

Analyzing the deep representations and their emergent structures gives insight into the generalization ability of deep features \cite{morcos2018importance} as well as the feature transferability across different tasks \cite{yosinski2014transferable}.
But current efforts on interpreting deep representations mainly focus on discriminative models \cite{zhou2014object,gonzalez2018semantic,zeiler2014visualizing,agrawal2014analyzing,bau2017network}.
Recent advance of Generative Adversarial Networks (GANs) \cite{gan,pggan,stylegan,biggan} is capable of transforming random noises into high-quality images, however, the nature of the learned generative representations and how a photo-realistic image is being composed over different layers of the generator in GAN remain much less explored.

It is known that the internal units of Convolutional Neural Networks (CNNs) emerge as object detectors when trained to categorize scenes \cite{zhou2014object}.
Representing and detecting objects most informative to a specific category provides an ideal solution for classifying scenes, such as sofa and TV are representative of the living room while bed and lamp are of the bedroom.
However, synthesizing a scene requires far more knowledge for the deep generative models to learn.
Specifically, in order to draw highly-diverse scene images, like our humans the deep representations might be required to not only learn to generate every individual object relevant to a specific scene category, but also decide the underlying room layout as well as render various scene attributes, \emph{e.g.}, the lighting condition and color scheme.
Recent work on interpreting GANs \cite{gandissection} visualized that the internal filters at intermediate layers are specialized for generating some certain objects, but studying scene synthesis from object aspect only is far from fully understanding how GAN is able to compose a photo-realistic image, which contains multiple variation factors from layout level, category level, to attribute level.
The original StyleGAN work \cite{stylegan} pointed out that the layer-wise latent codes actually control the synthesis from coarse to fine, but how these variation factors are composed together and how to quantify such semantic information remain unknown. 
Differently, this work gives a much deeper interpretation on the hierarchical generative representations in the sense that we match these layer-wise variation factors with human-understandable scene variations at multiple abstraction levels, including \emph{layout}, \emph{categorical object}, \emph{attribute}, and \emph{color scheme}.
Figure \ref{fig:teaser} shows the manipulation results at such various levels when the corresponding layers are identified correctly.

Starting with the state-of-the-art StyleGAN models \cite{stylegan} as the example, we reveal that highly-structured semantic hierarchy emerges from the deep generative representations with layer-wise stochasticity trained for synthesizing scenes, even without any external supervision.
Layer-wise representations are first probed with a broad set of visual concepts at different abstraction levels.
By quantifying the causality between the layer-wise activations and the semantics occurring in the output image, we are able to identify the most relevant variation factors across different layers of a GAN model with layer-wise latent codes: the early layers specify the spatial layout, the middle layers compose the category-guided objects, and the later layers render the attributes and color scheme of the entire scene.
We further show that identifying such a set of manipulatable variation factors from layouts, objects, to scene attributes and color schemes facilitates the semantic image manipulation (as shown in Fig.\ref{fig:teaser}) with a large diversity.
The proposed manipulation technique is further generalized to other GANs such as BigGAN \cite{biggan} and ProgressiveGAN \cite{pggan}.

\begin{figure*}[t]
  \centering
  \includegraphics[width=1.0\textwidth]{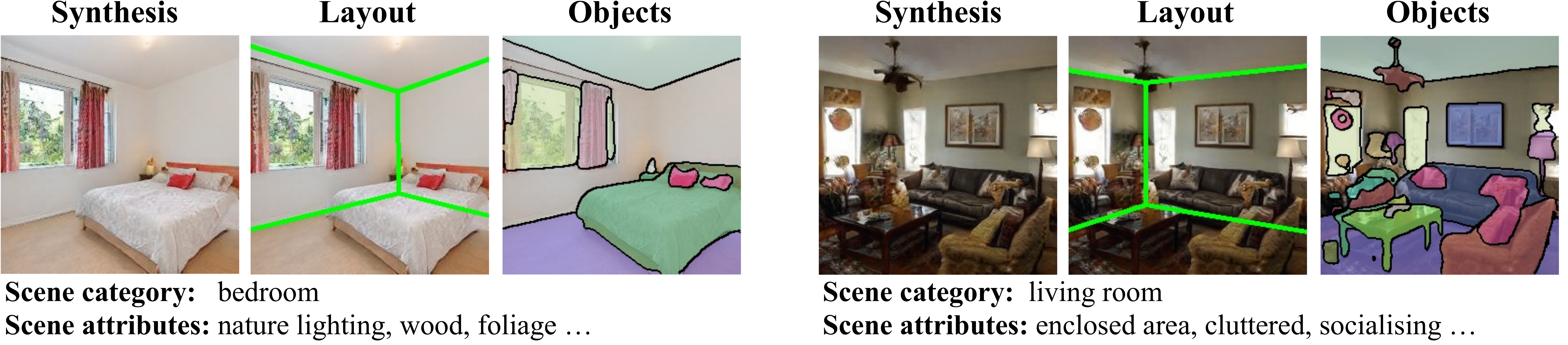}
  \captionsetup{font=small}
  \caption{
    Multiple levels of semantics extracted from two synthesized scenes.
  }
  \label{fig:semantics_example}
\end{figure*}

% !TEX root = main.tex

\section{Related Work}

\noindent\textbf{Deep representations from classifying images.}
Many attempts have been made to study the internal representations of CNNs trained for classification tasks.
\citet{zhou2014object} analyzed hidden units by simplifying the input image to see which context region gives the highest response, \citet{simonyan2013deep} applied the back-propagation technique to compute the image-specific class saliency map, \citet{bau2017network} interpreted the hidden representations via the aid of segmentation mask, \citet{alain2016understanding} trained independent linear probes to analyze the information separability among different layers.
There are also some studies transferring the features of CNNs to verify how learned representations fit with different datasets or tasks \cite{yosinski2014transferable,agrawal2014analyzing}.
In addition, reversing the feature extraction process by mapping a given representation back to image space \cite{zeiler2014visualizing,nguyen2016synthesizing,mahendran2015understanding} also gives insight into what CNNs actually learn to distinguish different categories.
However, these interpretation techniques developed for classification networks cannot be directly applied for generative models.

\noindent\textbf{Deep representations from synthesizing images.}
Generative Adversarial Networks (GANs) \cite{gan} advance the image synthesis significantly.
Some recent models \cite{pggan,biggan,stylegan} are able to generate photo-realistic faces, objects, and scenes, making GANs applicable to real-world image editing tasks, such as image manipulation \cite{shen2018faceid,xiao2018elegant,wang2018high,yao20183d}, image painting \cite{gandissection,park2019semantic}, and image style transfer \cite{zhu2017unpaired,choi2018stargan}.
Despite such a great success, it remains uncertain what GANs have actually learned to produce such diverse and realistic images.
\citet{dcgan} pointed out the vector arithmetic phenomenon in the underlying latent space of GAN, however, discovering what kinds of semantics exist inside a well-trained model and how these semantics are structured to compose high-quality images are still unsolved.
A very recent work \cite{gandissection} analyzed the individual units of the generator in GAN and found that they learn to synthesize informative visual contents such as objects and textures spontaneously.
Besides, concurrent work \cite{steerability,ganalyze} also explored the steerability and boosts the memorability of GANs via the learned semantics respectively.
Unlike them, our work quantitatively explores the emergence of multi-level semantics inside the layer-wise generative representations.

\noindent\textbf{Scene manipulation and editing.}
Previous efforts were also made to edit scene images.
\citet{laffont2014transient} defined 40 transient attributes and managed to transfer the appearance of a similar scene to the image for editing.
\citet{cheng2014imagespirit} proposed verbal guided image parsing to recognize and manipulate the objects in indoor scenes.
\citet{karacan2016learning} learned a conditional GAN to synthesize outdoor scenes based on pre-defined layout and attributes.
Some other work \cite{liao2017visual,zhu2017unpaired,isola2017image,luan2017deep} studied image-to-image translation and can be used to transfer the style of one scene to another.
Different from them, we achieve scene manipulation by interpreting the hierarchical semantics emerging from the generative representations of well-trained GANs.
Besides image editing, such interpretation also gives us a better insight on how generative models are able to produce photo-realistic synthesis.

% !TEX root = main.tex

\begin{figure}[t]
  \centering
  \includegraphics[width=0.48\textwidth]{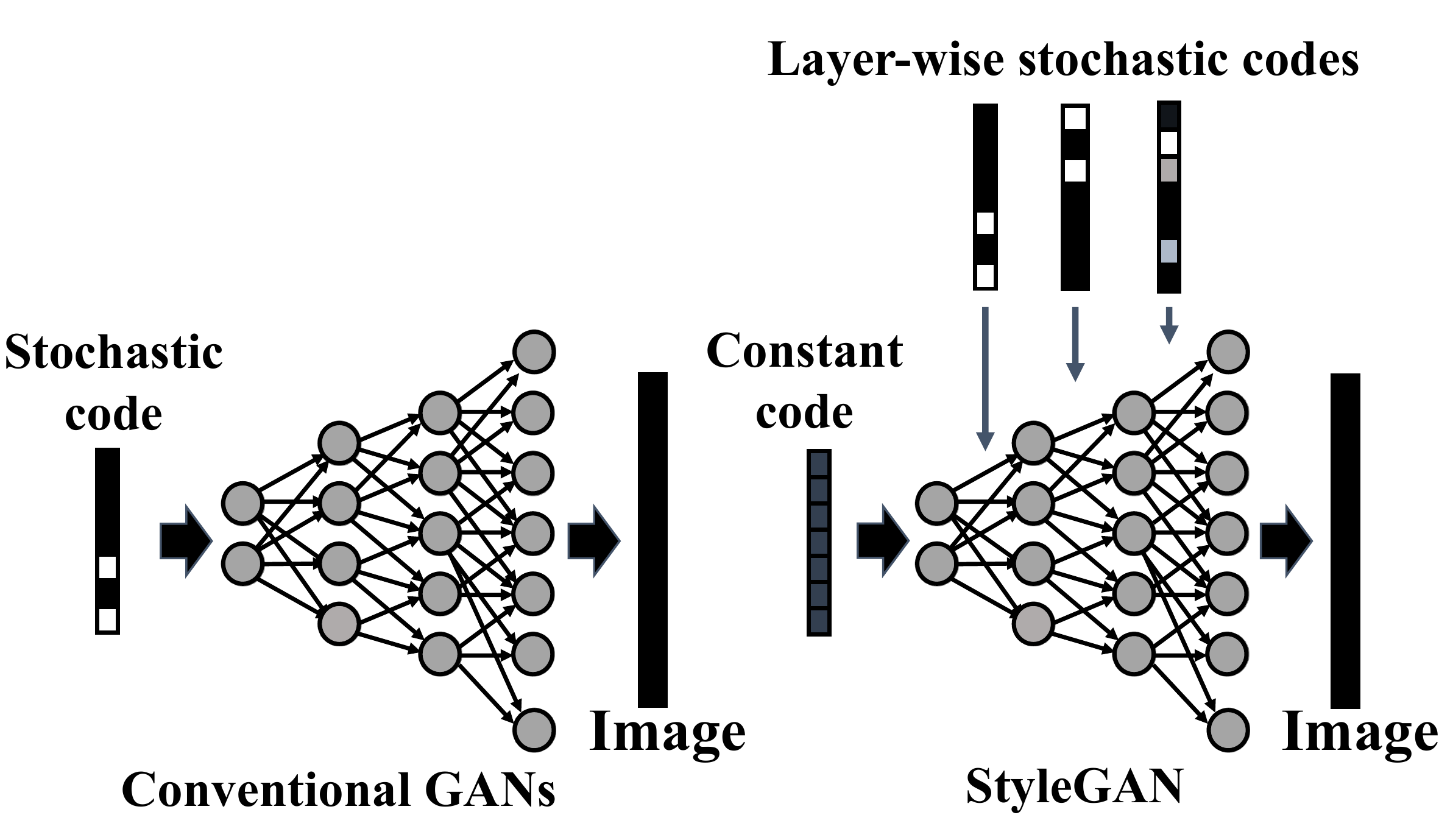}
  \captionsetup{font=small}
  \caption{
    Comparison between the conventional generator structure where the latent code is only fed into the very first layer and the generator in state-of-the-art GANs (\emph{e.g.}, StyleGAN \cite{stylegan} and BigGAN \cite{biggan}) which introduce layer-wise stochasticity by feeding latent codes to all convolutional layers.
  }
  \label{fig:laywerwise_stochasticity}
\end{figure}

\section{Variation Factors in Generative Representations}\label{sec:variation-factor-vs-generative-representation}

\subsection{Multi-Level Variation Factors for Scene Synthesis}\label{subsec:multi-level-variation-factors}
Imagine an artist drawing a picture of the living room.
The very first step, before drawing every single object, is to choose a perspective and set up the layout of the room.
After the spatial structure is set, the next step is to add objects that typically occur in a living room, such as a sofa and TV.
Finally, the artist will refine the details of the picture with specified decoration styles, \emph{e.g.}, warm or cold, natural lighting or indoor lighting.
The above process reflects how a human draws a scene by interpreting it from multiple abstraction levels. Meanwhile, given a scene image, we are able to extract multiple levels of attributes, as shown in Fig.\ref{fig:semantics_example}.
As a comparison, generative models such as GANs follow a completely end-to-end training manner for synthesizing scenes, without any prior knowledge about the drawing techniques and relevant concepts.
Even so, the trained GANs are able to produce photo-realistic scenes, which makes us wonder if the GANs have mastered any human-understandable drawing knowledge as well as the variation factors of scenes spontaneously.

\begin{figure*}[t!]
  \centering
  \includegraphics[width=1.0\textwidth]{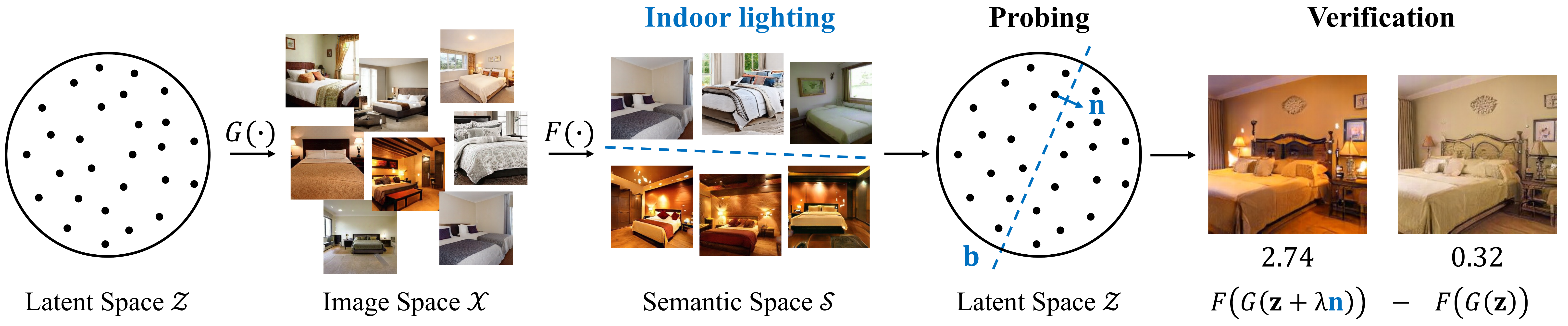}
  \captionsetup{font=small}
  \caption{
    Method for identifying the emergent variation factors in generative representation.
    By deploying a broad set of \emph{off-the-shelf} image classifiers as scoring functions, $F(\cdot)$, we are able to assign a synthesized image with semantic scores corresponding to each candidate variation factor.
    For a particular concept, we learn a decision boundary in the latent space by considering it as a binary classification task.
    Then we move the sampled latent code towards the boundary to see how the semantic varies in the synthesis, and use a re-scoring technique to quantitatively verify the emergence of the target concept.
  }
  \label{fig:framework}

\end{figure*}

\subsection{Layer-wise Generative Representations}\label{subsec:layer-wise-generative-representations}
In general, existing generative models take a randomly sampled latent code as input and output a image synthsis as real as possible.
Such a one-on-one mapping from latent codes to synthesized images is very similar to the feature extraction process in discrimination model.
Accordingly, in this work, we treat the input latent code as the \emph{generative representation} which will uniquely determine the appearance and properties of the output scene.
On the other hand, the recent state-of-the-art GAN models (\emph{e.g.}, StyleGAN \cite{stylegan} and BigGAN \cite{biggan}) introduce layer-wise stochasticity to improve the training stability and synthesis quality.
As shown in Fig.\ref{fig:laywerwise_stochasticity}, compared to the conventional generator which only takes the latent code as the input of the first layer, the improved generator with layer-wise stochasticity take random latent codes into all the layers.
We therefore treat them as layer-wise generative representations.
It is worth mentioning that more and more latest GAN models inherit the design of using layer-wise latent codes to achieve better generation quality, such as SinGAN \cite{shaham2019singan} and HoloGAN \cite{nguyen2019hologan}.

To explore how GANs are able to produce high-quality scene synthesis by learning multi-level variation factors as well as what role the generative representation of each layer plays in this process, this work aims at establishing the relationship between the variation factors and the generative representations.
\citet{stylegan} has already pointed out that the design of layer-wise stochasticity actually controls the synthesis from coarse to fine, however, what ``coarse'' and ``fine'' actually refer to still remains uncertain.
Differently, to align the variation factors with human perception, we separate them into four abstraction levels, including \emph{layout}, \emph{categorical objects}, \emph{scene attributes}, and \emph{color scheme}.
We further propose a framework in Sec.\ref{sec:framework} to quantify the causality between the input generative representations and the output variation factors.
We surprisingly find that GAN synthesizes a scene in a manner highly consistent with human.
Over all convolutional layers, GAN manages to compose these multi-level abstractions hierarchically.
In particular, GAN constructs the spatial layout at the early stage, synthesizes category-specified objects at the middle stage, and renders the scene attribute and color scheme at the later stage.

\section{Identifying the Emergent Variation Factors}\label{sec:framework}
As described in Sec.\ref{sec:variation-factor-vs-generative-representation}, we target at interpreting the latent semantics learned by scene synthesis models from four different abstraction levels.
Previous efforts on several scene understanding databases \cite{zhou2017places, sundatabase, laffont2014transient, sceneattribute} enable a series of classifiers to predict scene attributes and categories.
Besides, we also employ sevearl off-the-shelf classifiers focusing on layout detection \cite{layout} and semantic segmentation \cite{xiao2018unified} to help analyze the synthesized scene images.
Specially, given an image, we are able to use these classifiers to get the response scores with respect to various semantics.
However, only predicting the semantic labels is far from identifying the variation factors that GANs have captured from the training data.
More concretely, among all the multi-level candidate concepts, not all of them are meaningful to a particular scene synthesis model.
For instance, ``indoor lighting'' will never happen in outdoor scenes such as bridge and tower, which ``enclosed area'' is always true for indoor scenes such as bedroom and kitchen.
Accordingly, we come up with a method to quantitatively identify the most relevant and manipulatable variation factors that emerge inside the learned generative representation.
Fig.\ref{fig:framework} illustrates the identification process which consists of two steps, \emph{i.e.}, probing (Sec.\ref{subsec:probing}) and verification (Sec.\ref{subsec:verifying}).
Such identification enables the diverse scene manipulation (Sec.\ref{subsec:manipulating}).

\subsection{Probing Latent Space}\label{subsec:probing}
The generator of GAN, $G(\cdot)$, typically learns the mapping from latent space $\Z$ to image space $\X$.
Latent vectors $\z\in\Z$ can be considered as the generative representation learned by GAN.
To study the emergence of variation factors inside $\Z$, we need to first extract semantic information from $\z$, which is not trivial.
To solve this problem, we employ synthesized image, $\x = G(\z)$, as an intermediate step and use a broad set of \emph{off-the-shelf} image classifiers to help assign semantic scores for each sampled latent code $\z$.
Taking ``indoor lighting'' as an example, the scene attribute classifier is able to output the probability of how an input image looks like having indoor lighting, which we use as the semantic score.
Recall that we divide scene representation into layout, object (category), and attribute levels, we introduce layout estimator, scene category recognizer, and attribute classifier to predict semantic scores from these abstraction levels respectively, forming a hierarchical semantic space $\S$.
After establishing the one-on-one mapping from latent space $\Z$ to sematic space $\S$, we search the decision boundary for each concept by treating it as a bi-classification problem, as shown in Fig.\ref{fig:framework}.
Here, taking ``indoor lighting'' as an instance, the boundary separates the latent space $\Z$ to two sets, \emph{i.e.}, presence or absence of indoor lighting.

\subsection{Verifying Manipulatable Variation Factors}\label{subsec:verifying}
After probing the latent space with a broad set of candidate concepts, we still need to figure out which ones are most relevant to the generative model acting as the variation factors.
The key issue is how to define ``relevance'', or say, how to verify whether the learned representation has already encoded a particular variation factor.
We argue that if the target concept is manipulatable from latent space perspective (\emph{e.g.}, change the indoor lighting status of the synthesized image via simply varying the latent code), the GAN model is able to capture such variation factors during the training process.

As mentioned above, we have already got separation boundaries for each candidate.
Let $\{\n_i\}_{i=1}^C$ denote the normal vectors of these boundaries, where $C$ is the total number of candidates.
For a certain boundary, if we move a latent code $\z$ along its normal direction (positive), the semantic score should also increase correspondingly.
Therefore, we propose to re-score the varied latent code to \emph{quantify} how a variation factor is relevant to the target model for analysis.
As shown in Fig.\ref{fig:framework}, this process can be formulated as
\begin{align}
  \Delta s_i = \frac{1}{K} \sum_{k=1}^{K} \max\Big(F_i\big(G(\z^k+\lambda\n_i)\big) - F_i\big(G(\z^k)\big),0\Big), \label{eq:rescore}
\end{align}
where $\frac{1}{K} \sum_{k=1}^{K}$ stands for the average of $K$ samples to make the metric more accurate.
$\lambda$ is a fixed moving step.
To make this metric comparable among all candidates, all normal vectors $\{\n_i\}_{i=1}^C$ are normalized to fixed norm 1 and $\lambda$ is set as 2.
With this re-scoring technique, we can easily rank the score $\Delta s_i$ among all $C$ concepts to retrieve the most relevant latent variation factors.

\begin{figure}[t]
  \centering
  \includegraphics[width=0.48\textwidth]{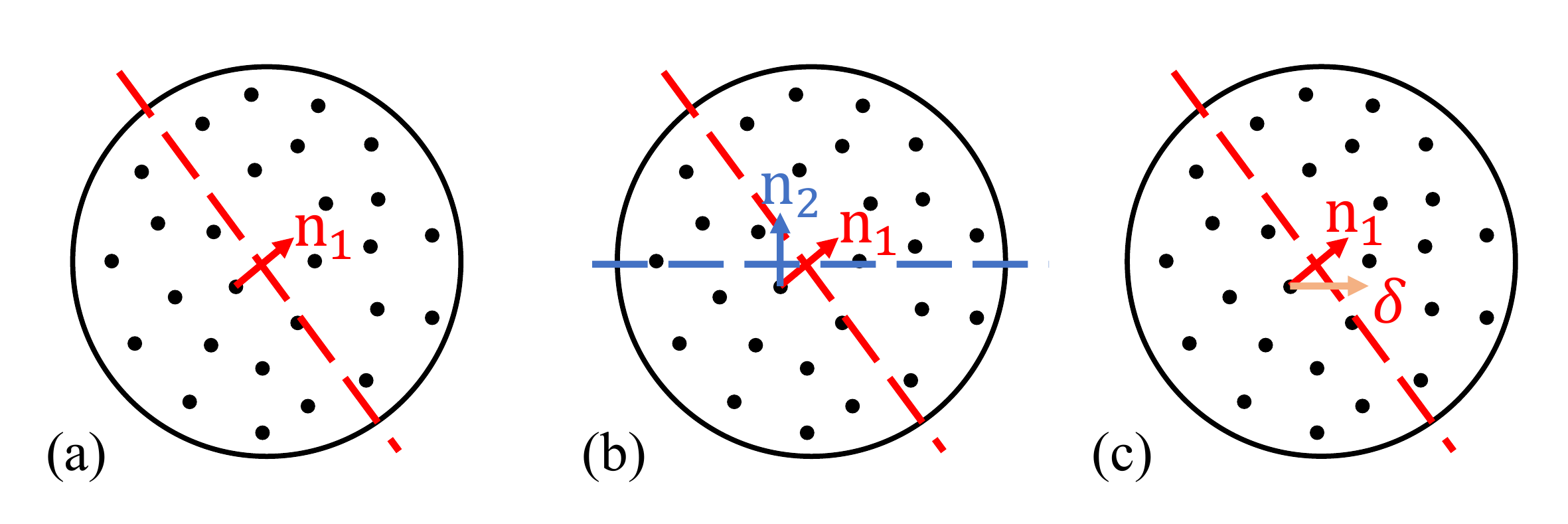}
  \captionsetup{font=small}
  \caption{
    Three types of manipulation: (a) \emph{Independent} manipulation; (b) \emph{Joint} manipulation; (c) \emph{Jittering} manipulation.
  }
  \label{fig:diverse_manipulation}
\end{figure}

\subsection{Manipulation with Diversity}\label{subsec:manipulating}
%1
After identifying the semantic variation facors, we propose several ways to futher manipulate images. Figure \ref{fig:diverse_manipulation} shows three types of scene manipulation.
A simple and straightforward way, named \emph{independent} manipulation, is to push the code $\z$ along the normal vector $\n_i$ of a certain semantic at the step length of $\lambda$.
The manipulated code $\z' \leftarrow \z +\lambda \n $ is then fed into the generator to produce the new image.
A second way of manipulation enables scene editing with respect to more than one variation factor jointly.
We call it \emph{joint} manipulation.
Taking two variation factors (with normal vector $\n_1$ and $\n_2$) as an example, the original code $\z$ is moved along the two directions simultaneously as $\z' \leftarrow \z+ \lambda_1 \n_1 + \lambda_2 \n_2$.
Here $\lambda_1$ and $\lambda_2$ are step parameters which control the strength of manipulation corresponding to these two semantic respectively.
Since such two manipulation methods enable more precise control from multiple abstraction levels, 
we also introduce randomness into the manipulation process to increase the diversity, namely \emph{jittering} manipulation.
The key idea is to slightly pertub the manipulation direction with a randomly sampled noise $\delta \sim \mathcal{N}(0, 1)$.
It can be then formulated as $\z' \leftarrow \z+ \lambda \n + \delta$.

% !TEX root = main.tex

% Figure: Training Samples
\begin{figure*}[t]
  \centering
  \includegraphics[width=1.0\textwidth]{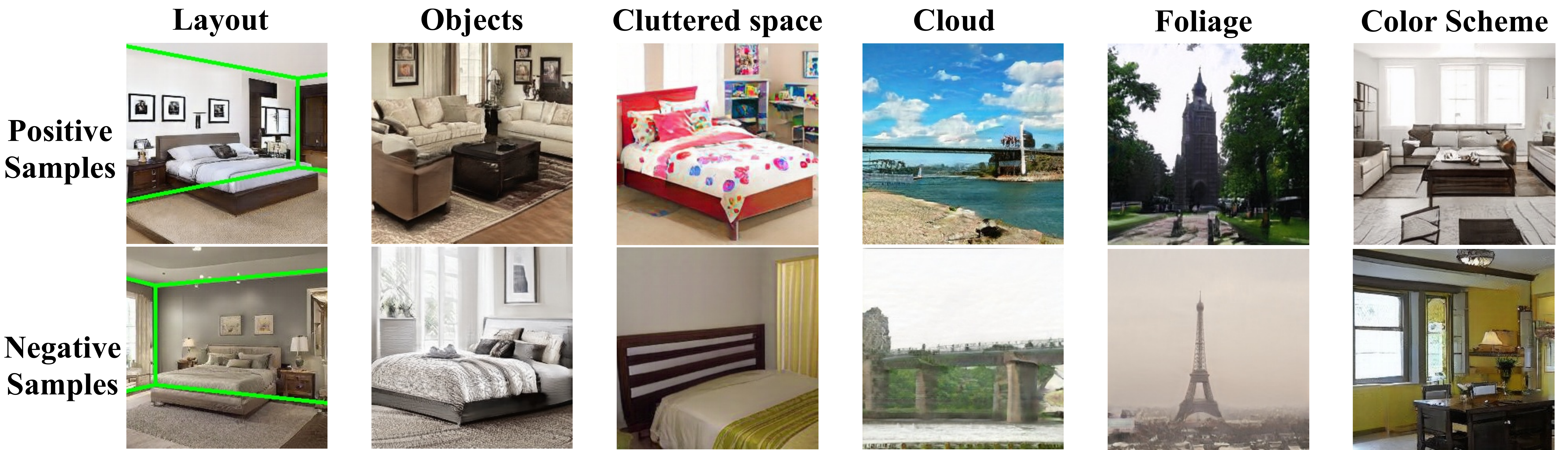}
  \captionsetup{font=small}
  \caption{
    Diverse generated samples for training decision boundary with respect to layout, objects (category), scene attributes and color scheme.
  }
  \label{fig:training-samples}
\end{figure*}

\begin{figure}[t]
  \centering
  \includegraphics[width=0.4\textwidth]{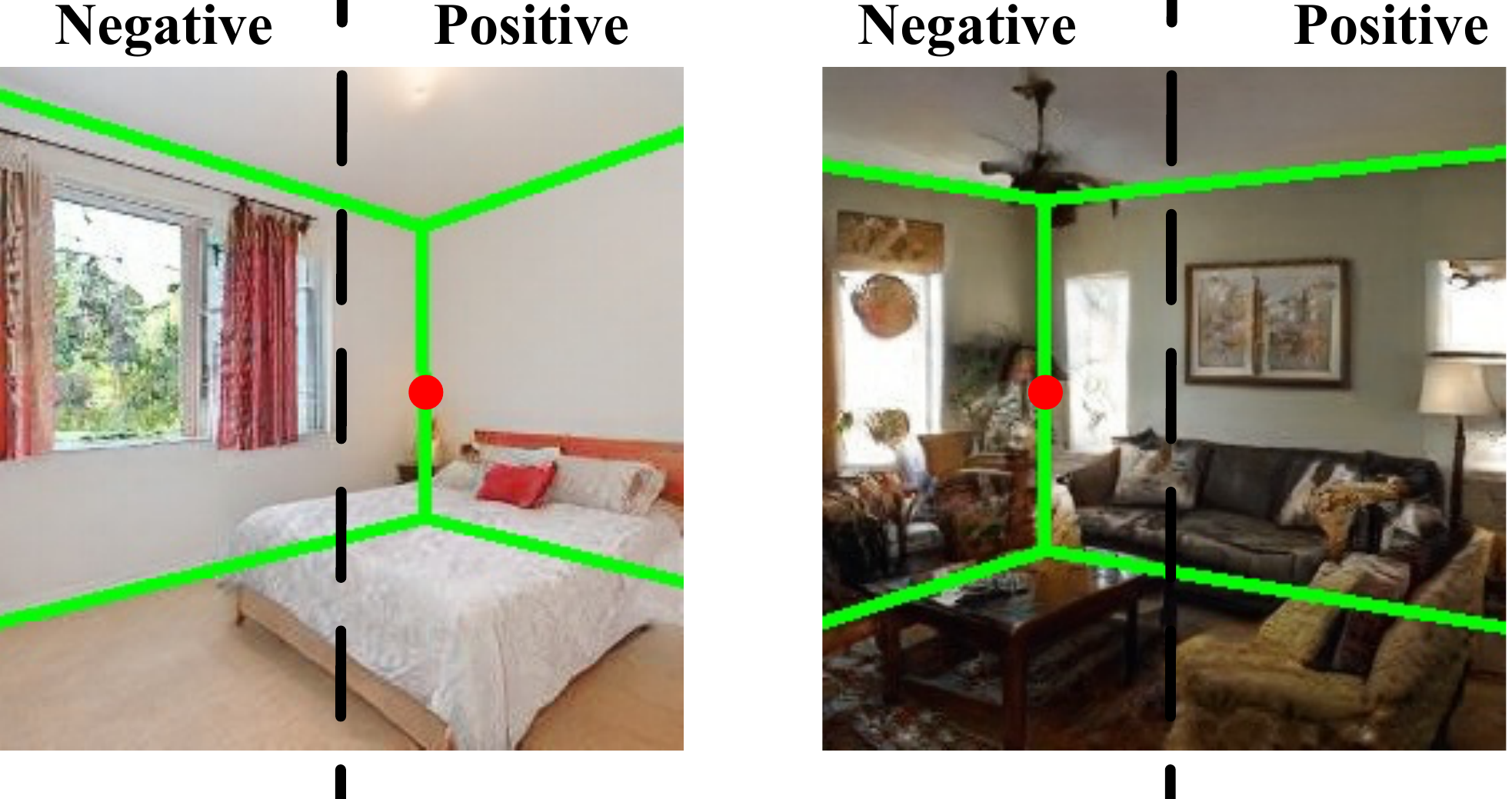}
  \captionsetup{font=small}
  \caption{
    The definition of layout for indoor scenes.
    \textcolor{green}{\textbf{Green}} lines represent for the outline prediction from the layout estimator.
    The dashed line indicates the horizontal center, and the \textcolor{red}{\textbf{red}} point is the center point of the intersection line of two walls.
    The relative position between the vertical line and the center point is used to split the dataset.
  }
  \label{fig:layout}
\end{figure}

\setlength{\tabcolsep}{5pt}
\begin{table}[t]
  \captionsetup{font=small}
  \caption{Description of the StyleGAN models trained on different categories. $\downarrow$ means the lower the better.}
  \label{tab:style-gan}
  \small\centering
  \begin{tabular}{cccc}
    \toprule
    Scene Category         & Indoor / Outdoor & Training Samples  & FID $\downarrow$ \\
    \midrule
    bedroom (official)     & Indoor           & 3M                & 2.65 \\
    living room            & Indoor           & 1.3M              & 5.16 \\
    kitchen                & Indoor           & 1M                & 5.06 \\
    % dining room          & Indoor           & 658K             & 25M               & 4.13 \\
    restaurant             & Indoor           & 626K              & 4.03 \\
    % classroom            & Indoor           & 168K             & 50M               & 9.75 \\
    % conference room      & Indoor           & 229K             & 50M               & 6.20 \\
    bridge                 & Outdoor          & 819K              & 6.42 \\
    church                 & Outdoor          & 126K              & 4.82 \\
    tower                  & Outdoor          & 708K              & 5.99 \\
    Mixed                  & Indoor           & 500K each         & 3.74 \\
    \bottomrule
  \end{tabular}
\end{table}

\section{Experiments}

In the generation process, the deep representation at each layer, especially for StyleGAN \cite{stylegan} and BigGAN \cite{biggan}, is actually directly derived from the projected latent code.
Therefore, we consider the latent code as the \emph{generative representation}, which may be slightly different from the conventional definition in the classification networks.
We conduct a detailed empirical analysis of the variation factors identified across the layers of the generators in GANs.
We show that the hierarchy of variation factors emerges in the deep generative representations as a result of learning to synthesize scenes. 

The experimental section is organized as follows: 
Sec.\ref{subsec:implementation-details} introduces our experimental details including generative models, training datasets and the \emph{off-the-shelf} classifiers we used.
Sec.\ref{subsec:hierarchical-semantics} contains the layer-wise analysis on the state-of-the-art StyleGAN model \cite{stylegan}, quantitatively and qualitatively verifying that the multi-level variation factors are encoded in the latent space.
In Sec.\ref{subsec:categorical-analysis}, we explore the question on how GANs represent categorical information such as bedroom \emph{v.s.} living room.
We reveal that GAN synthesizes the shared objects at some intermediate layers.
By controlling their activations only, we can easily overwrite the category of the output image, \emph{e.g.} turning bedroom into living room, while preserving its original layout and high-level attributes such as indoor lighting.
Sec.\ref{subsec:attribute-identification} further shows that our approach can faithfully identify the most relevant attributes associated with a particular scene, facilitating semantic scene manipulation.
Sec.\ref{subsec:ablation} conducts the ablation studies on re-scoring technique and layer-wise manipulation to show the effectiveness of our approach.

\subsection{Experimental Details}\label{subsec:implementation-details}

\noindent \textbf{Generator models.}
This work conducts experiments on state-of-the-art deep generative models for high-resolution scene synthesis, including StyleGAN \cite{stylegan}, BigGAN \cite{biggan}, and PGGAN \cite{pggan}.
Among them, PGGAN employs the conventional generator structure where the latent code is only fed into the very first layer.
Differently, StyleGAN and BigGAN introduce layer-wise stochasticity by feeding latent codes to all convolutional layers as shown in Fig.\ref{fig:laywerwise_stochasticity}.
And our layer-wise analysis sheds light on why it is effective.

\begin{figure*}[t!]
  \centering
  \includegraphics[width=0.95\textwidth]{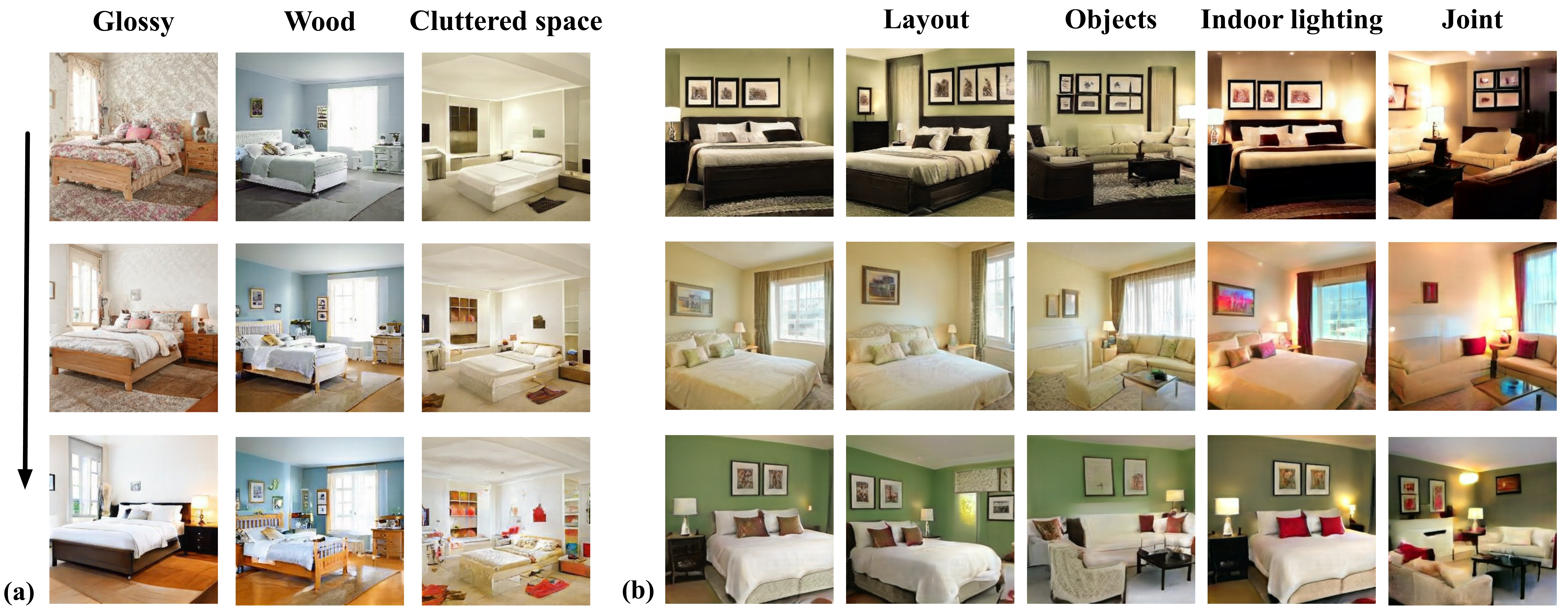}
  \captionsetup{font=small}
  \caption{
    (a) \emph{Independent} attribute manipulation results on Upper layers. The middle row is the source images. We are able to both decrease (top row) and increase (bottom row) the variation factors in the images.
    (b) \emph{Joint} manipulation results, where the \emph{layout}, \emph{objects} and \emph{attribute} are manipulated at proper layers. The first column indicates the source images, the middle three columns are the independently manipulated images.
  }
  \label{fig:layer-wise-manipulation}
\end{figure*}

\noindent \textbf{Scene categories.}
Among the mentioned generator models, PGGAN and StyleGAN are actually trained on LSUN dataset \citet{yu2015lsun} while BigGAN is trained on Places dataset \cite{zhou2017places}.
LSUN dataset consists of 7 indoor scene categories and 3 outdoor scene categories, and Places dataset contains 10 million images across 434 categories.
For PGGAN model, we use the officially released models\footnote{These PGGAN models can be found at \url{https://drive.google.com/open?id=15hvzxt_XxuokSmj0uO4xxMTMWVc0cIMU}.}, each of which is trained to synthesize scene within a particular category of LSUN dataset.
For StyleGAN, only one model related to scene synthesis (\emph{i.e.}, bedroom) is released\footnote{The StyleGAN model can be found at \url{https://drive.google.com/drive/folders/1MASQyN5m0voPcx7-9K0r5gObhvvPups7}.}.
For a more thorough analysis, we use the official implementation\footnote{The implementation of StyleGAN can be found at \url{https://github.com/NVlabs/stylegan}.} to train some additional models on other scene categories, including both indoor scenes (living room, kitchen, restaurant) and outdoor scenes (bridge, church, tower).
We also train a \emph{mixed} model on the combination of images from bedroom, living room, and dining room with the same implementation.
This model is specifically used for categorical analysis.
For each StyleGAN model, Tab.\ref{tab:style-gan} shows the category, the number of training samples, as well as the corresponding Fr\'echet inception distances (FID) \cite{fid} which can reflect the synthesis quality to some extent.
For BigGAN, we use the author's officially unofficial PyTorch BigGAN implementation\footnote{The implementation of BigGAN can be found at \url{https://github.com/ajbrock/BigGAN-PyTorch}.} to train a conditional generative model by taking category label as the constraint on Places dataset \cite{zhou2017places}.
The resolution of the scene images synthesized by all of the above models is $256\times256$.

\noindent \textbf{Semantic Classifiers.}
To extract semantic from synthesized images, we employ some \emph{off-the-shelf} image classifiers to assign these images with semantic scores from multiple abstraction levels, including \emph{layout}, \emph{category}, \emph{scene attribute}, and \emph{color scheme}.
Specifically, we use
(1) a \emph{layout estimator} \cite{layout}, which predicts the spatial structure of an indoor place,
(2) a \emph{scene category classifier} \cite{zhou2017places}, which classifies a scene image to 365 categories,
and (3) an \emph{attribute predictor} \cite{zhou2017places}, which predicts 102 pre-defined scene attributes in SUN attribute database \cite{sceneattribute}.
We also extract color scheme of a scene image through its hue histogram in HSV space.
Among them, the category classifier and attribute predictor can directly output the probability of how likely an image belongs to a certain category or how likely an image has a particular attribute.
As for the layout estimator, it only detects the outline structure of an indoor place, shown as the green line in Fig.\ref{fig:layout}.

% Figure: Layer-wise Analysis

\noindent \textbf{Semantic Probing and Verification.}
Given a well-trained GAN model for analysis, we first generate a collection of synthesized scene images by randomly sampling $N$ latent codes.
To ensure capturing all the potential variation factors, we set $N=500,000$.
We then use the aforementioned image classifiers to assign semantic scores for each visual concept.
It is worth noting that we use the relative position between image horizontal center and the intersection line of two walls to quantify layout, as shown in Fig.\ref{fig:layout}.
After that, for each candidate, we select $2,000$ images with the highest response as positive samples, and another $2,000$ with the lowest response as negative ones.
Fig.\ref{fig:training-samples} shows some examples, where the living room and bedroom are treated as positive and negative for scene category respectively.
We then train a linear SVM by treating it as a bi-classification problem (\emph{i.e.}, data is the sampled latent code while the label is binary indicating whether the target semantic appears in the corresponding synthesis or not) to get a linear decision boundary.
Finally, we re-generate $K=1,000$ samples for semantic verification as described in Sec.\ref{subsec:verifying}.

% Figure: Layer-wise Manipulation

\begin{figure}[t]
  \centering
  \includegraphics[width=0.48\textwidth]{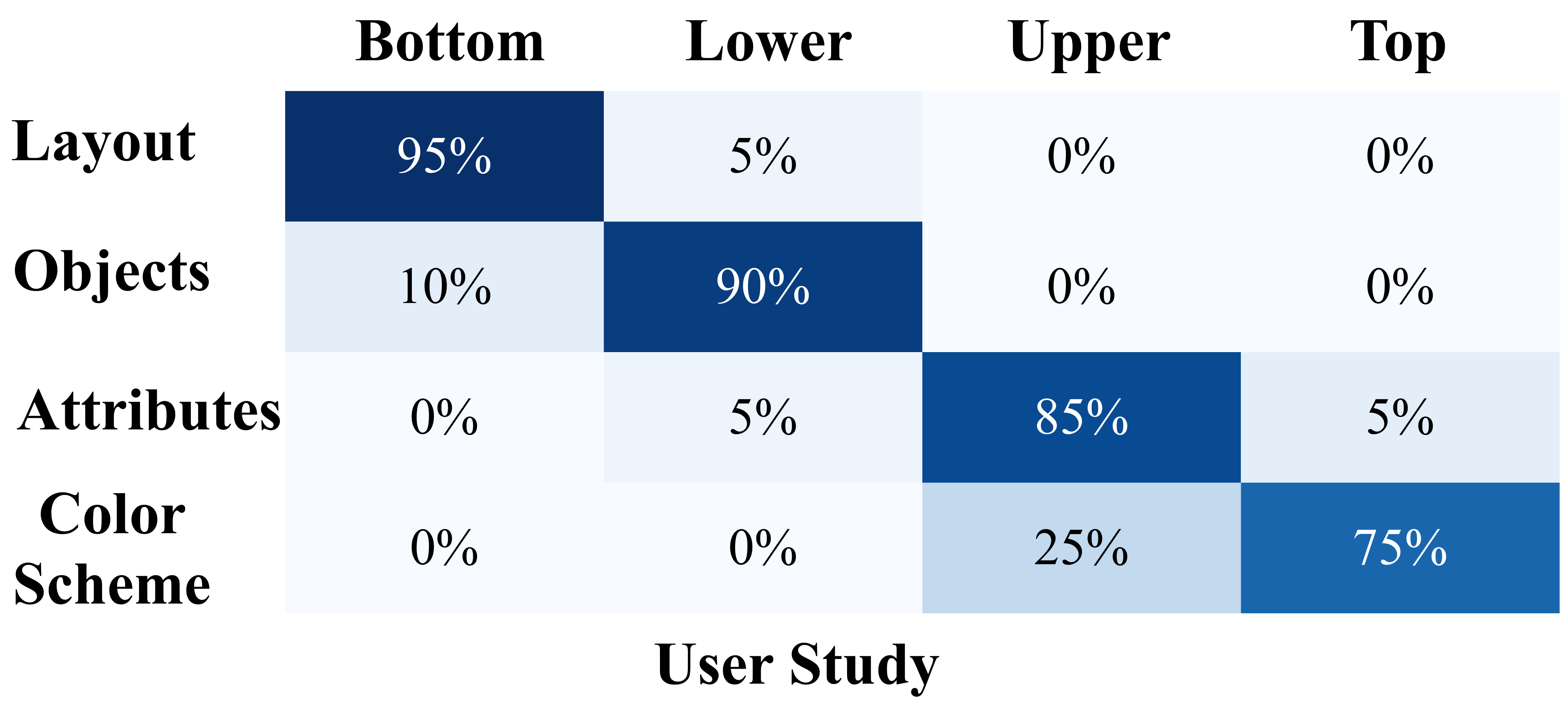}
  \captionsetup{font=small}
  \caption{
    User study on how different layers correspond to variation factors from different abstraction levels.
  }
  \label{fig:userstudy}
\end{figure}

\begin{figure}[t]
  \centering
  \includegraphics[width=0.48\textwidth]{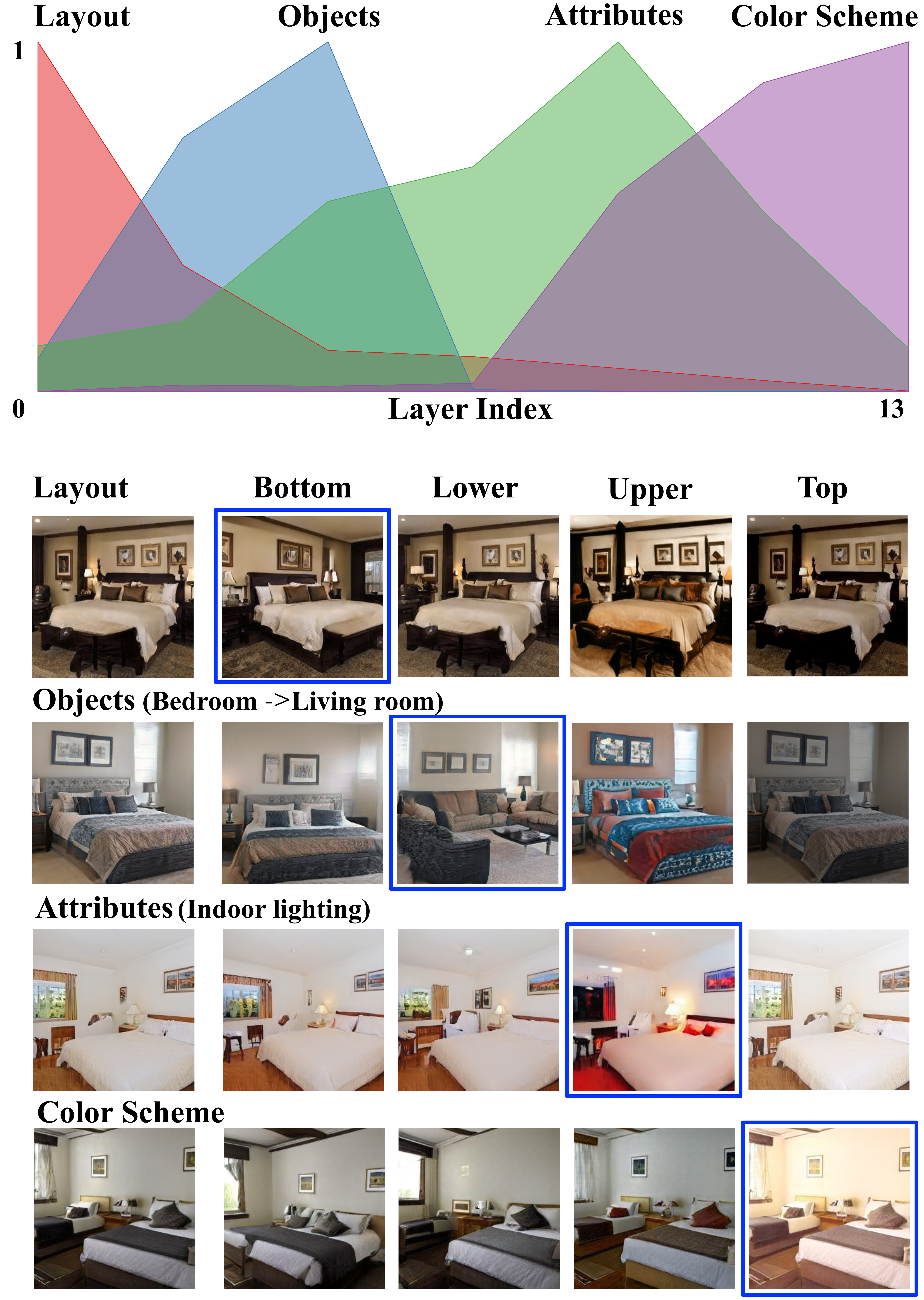}
  \captionsetup{font=small}
  \caption{
    Top: Four levels of visual abstractions emerge at different layers of StyleGAN. Vertical axis shows the normalized perturbation score $\Delta s_i$.
    Bottom: Layer-wise manipulation result. The first column is the original synthesized images, and the other columns are the manipulated images at layers from four different stages respectively. Blue boxes highlight the results from varying the latent code at the most proper layers for the target concept.
  }
  \label{fig:layer-wise-analysis}
\end{figure}

\begin{figure}[t]
  \centering
  \includegraphics[width=0.48\textwidth]{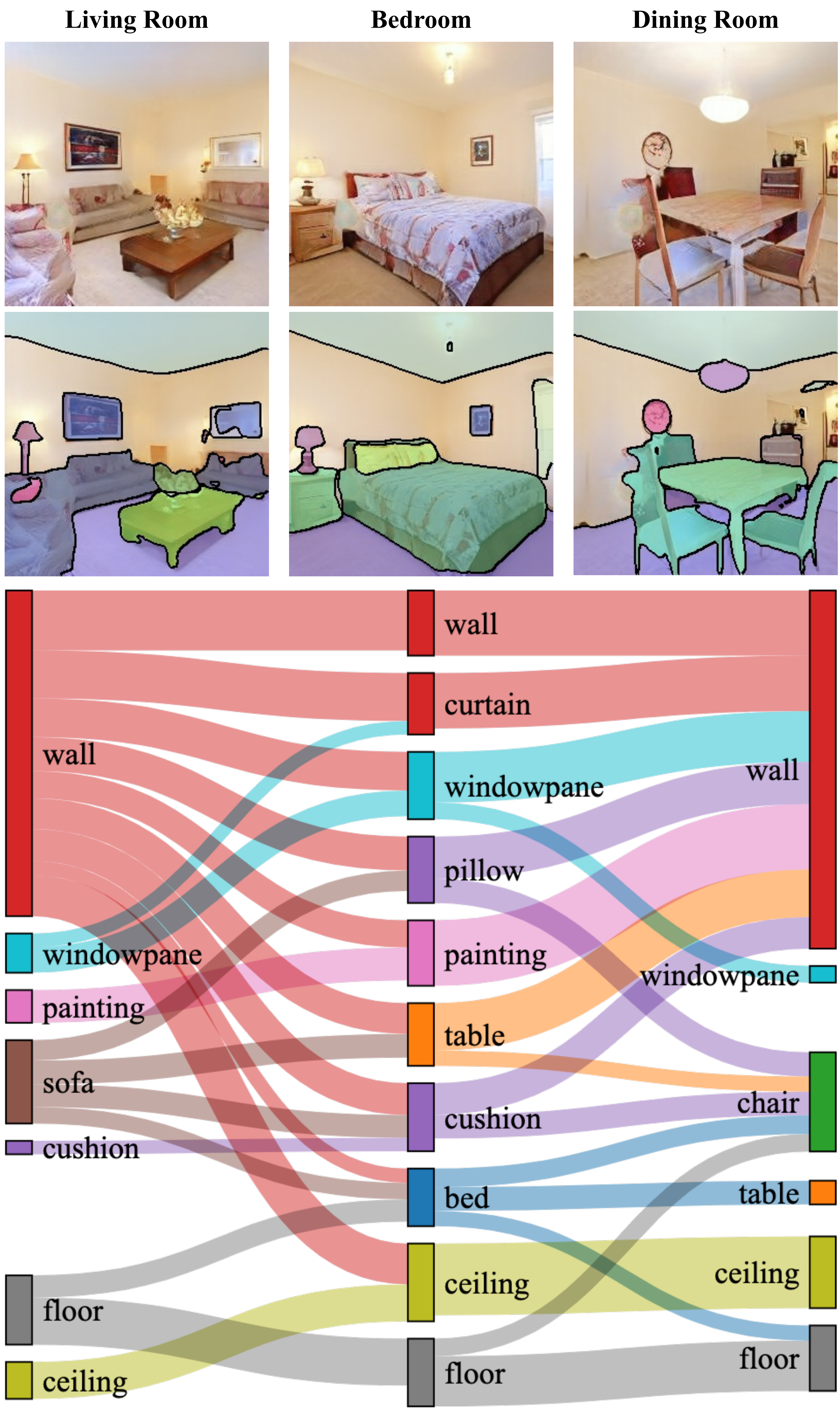}
  \captionsetup{font=small}
  \caption{
    Objects are transformed by GAN to represent different scene categories.
    On the top shows that the object segmentation mask varies when manipulating a living room to bedroom, and further to dining room.
    On the bottom visualizes the object mapping that appears during category transition, where pixels are counted only from object level instead of instance level.
    GAN is able to learn shared objects as well as the transformation of objects with similar appearance when trained to synthesize scene images from more than one category.
  }
  \label{fig:seg-example}
\end{figure}

\begin{figure*}[t]
  \centering
  \includegraphics[width=1.0\textwidth]{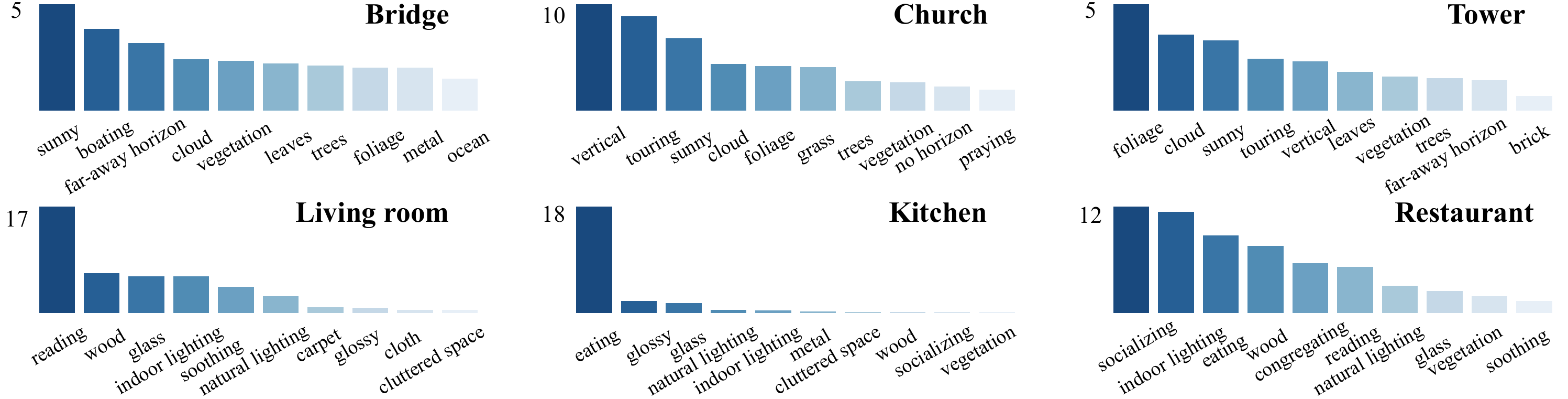}
  \captionsetup{font=small}
  \caption{
    Comparison of the top scene attributes identified in the generative representations learned by StyleGAN models for synthesizing different scenes.
    Vertical axis shows the perturbation score $\Delta s_i$.
  }
  \label{fig:scene-comparison-attribute}
\end{figure*}

\subsection{Emerging Semantic Hierarchy}\label{subsec:hierarchical-semantics}
Humans typically interpret a scene in a hierarchy of semantics, from its layout, underlying objects, to the detailed attributes and the color scheme.
Here the underlying objects refer to the set of objects most relevant to a specific category.
This section shows that GAN composes a scene over the layers in a similar way with human perception.
To enable analysis on layout and object, we take the \emph{mixed} StyleGAN model trained on indoor scenes as the target model.
StyleGAN \cite{stylegan} learns a more disentangled latent space $\W$ on top of the conventional latent space $\Z$ and feeds the latent code $\w\in\W$ to each convolutional layer with different transformations instead of only feeding it to the first layer.
Specifically, for $\ell$-th layer, $\w$ is linearly transformed to layer-wise transformed latent code $\y^{(\ell)}$ with $\y^{(\ell)} = \A^{(\ell)}\w + \b^{(\ell)}$, where $\A^{(\ell)}$, $\b^{(\ell)}$ are the weight and bias for style transformation respectively.
We thus perform layer-wise analysis by studying $\y^{(\ell)}$ instead of $\z$ in Eq.(\ref{eq:rescore}).

To quantify the importance of each layer with respect to each variation factor,  we use the re-scoring technique to identify the causality between the layer-wise generative representation $\y^{(\ell)}$ and the semantic emergence.
The normalized score in the top Fig.\ref{fig:layer-wise-analysis} shows that the layers of the generator in GAN are specialized to compose semantics in a hierarchical manner: the bottom layers determine the layout, the lower layers and upper layers control category-level and attribute-level variations respectively, while color scheme is mostly rendered at the top.
This is consistent with human perception.
In StyleGAN model that is trained to produce $256\times256$ scene images, there are totally 14 convolutional layers.
According to our experimental results, \emph{layout}, \emph{object (category)}, \emph{attribute}, \emph{color scheme} correspond to \emph{bottom}, \emph{lower}, \emph{upper}, and \emph{top} layers respectively, which are actually $\left[0,2\right)$, $\left[2,6\right)$, $\left[6,12\right)$ and $\left[12, 14\right)$ layers.

To visually inspect the identified variation factors, we move the latent vector along the boundaries at different layers to show how the synthesis varies correspondingly.
For example, given a boundary in regards to room layout, we vary the latent code towards the normal direction at bottom, lower, upper, and top layers respectively.
The bottom of Fig.\ref{fig:layer-wise-analysis} shows the qualitative results for several concepts.
We see that the emerged variation factors follow a highly-structured semantic hierarchy, \emph{e.g.}, layout can be best controlled at the early stage while color scheme can only be changed at the final stage.
Besides, varying latent code at the inappropriate layers may also change the image content, but the changing might be inconsistent with the desired output.
For example, in the second row, modulating the code at bottom layers for category only leads to a random change in the scene viewpoint.

To better evaluate the manipulability across layers, we conduct a user study.
We first generate 500 samples and manipulate them with respect to several concepts on different layers.
For each concept, 20 users are asked to choose the most appropriate layers for manipulation.
Fig.\ref{fig:userstudy} shows the user study results, where most people think bottom layers best align with layout, lower layers control scene category, \emph{etc}.
This is consistent with our observations in Fig.\ref{fig:layer-wise-analysis}.
It suggests that hierarchical variation factors emerge inside the generative representation for synthesizing scenes.
and that our re-scoring method indeed helps identify the variation factors from a broad set of semantics.

Identifying the semantic hierarchy and the variation factors across layers facilitates semantic scene manipulation.
We can simply push the latent code toward the boundary of the desired attribute at the appropriate layer.
Fig.\ref{fig:layer-wise-manipulation}(a) shows that we can change the decoration style (crude to glossy), the material of furniture (cloth to wood), or even the cleanliness (tidy to cluttered) respectively.
Furthermore, we can jointly manipulate hierarchical variation factors.
In Fig.\ref{fig:layer-wise-manipulation}(b) we simultaneously change the room layout (rotating viewpoint) at early layers, scene category (converting bedroom to living room) at middle layers, and scene attribute (increasing indoor lighting) at later layers.

\begin{figure*}[!t]
  \centering
  \includegraphics[width=1\textwidth]{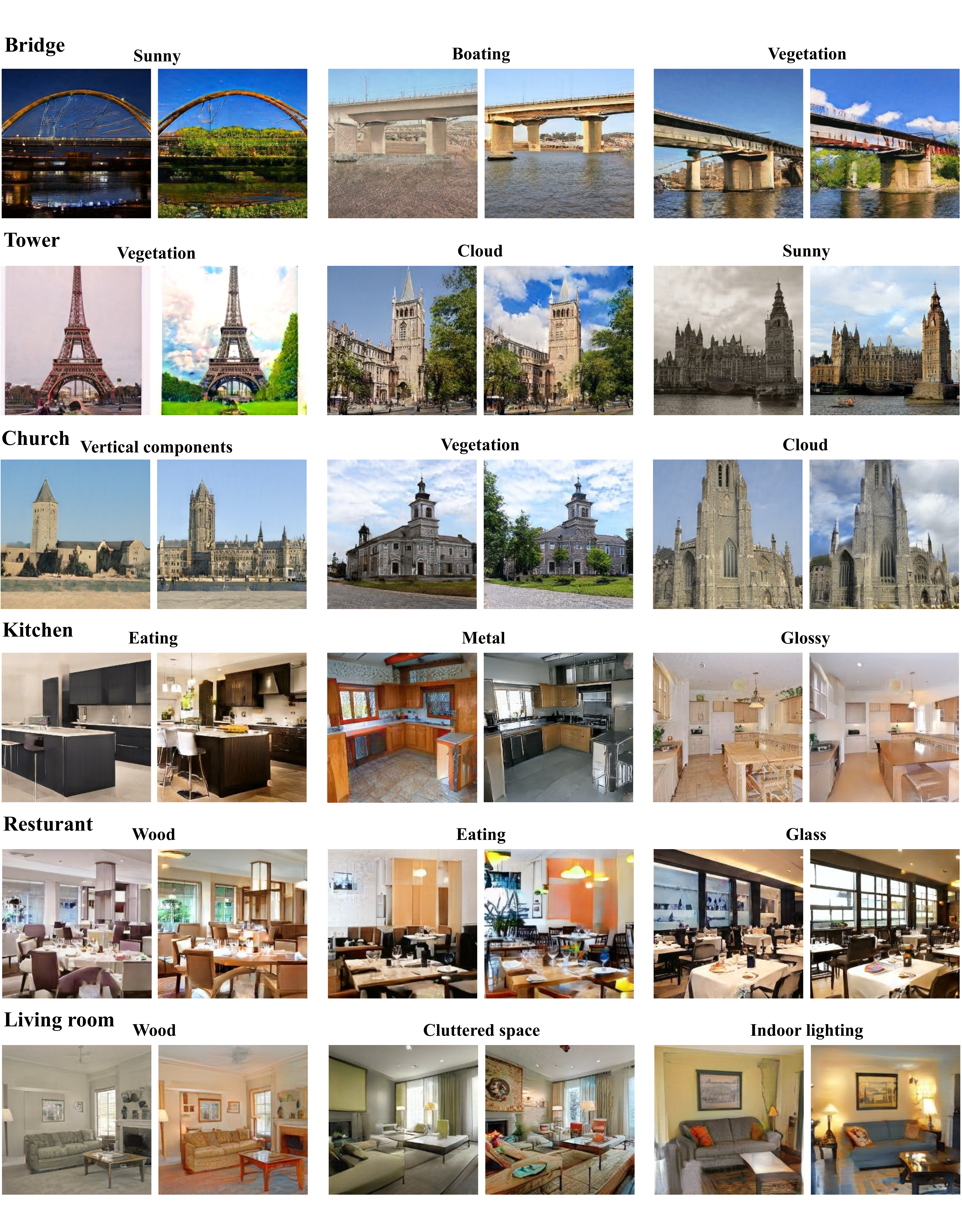}
  \captionsetup{font=small}
  \caption{
    \emph{Independent} manipulation results on StyleGAN models trained for synthesizing indoor and outdoor scenes. In each pair of images, the first is the original synthesized sample and the second is the one after the manipulation of a certain semantics.
  }
  \label{fig:outdoor_manipulation}
\end{figure*}

\begin{figure*}[t]
  \centering
  \includegraphics[width=1.0\textwidth]{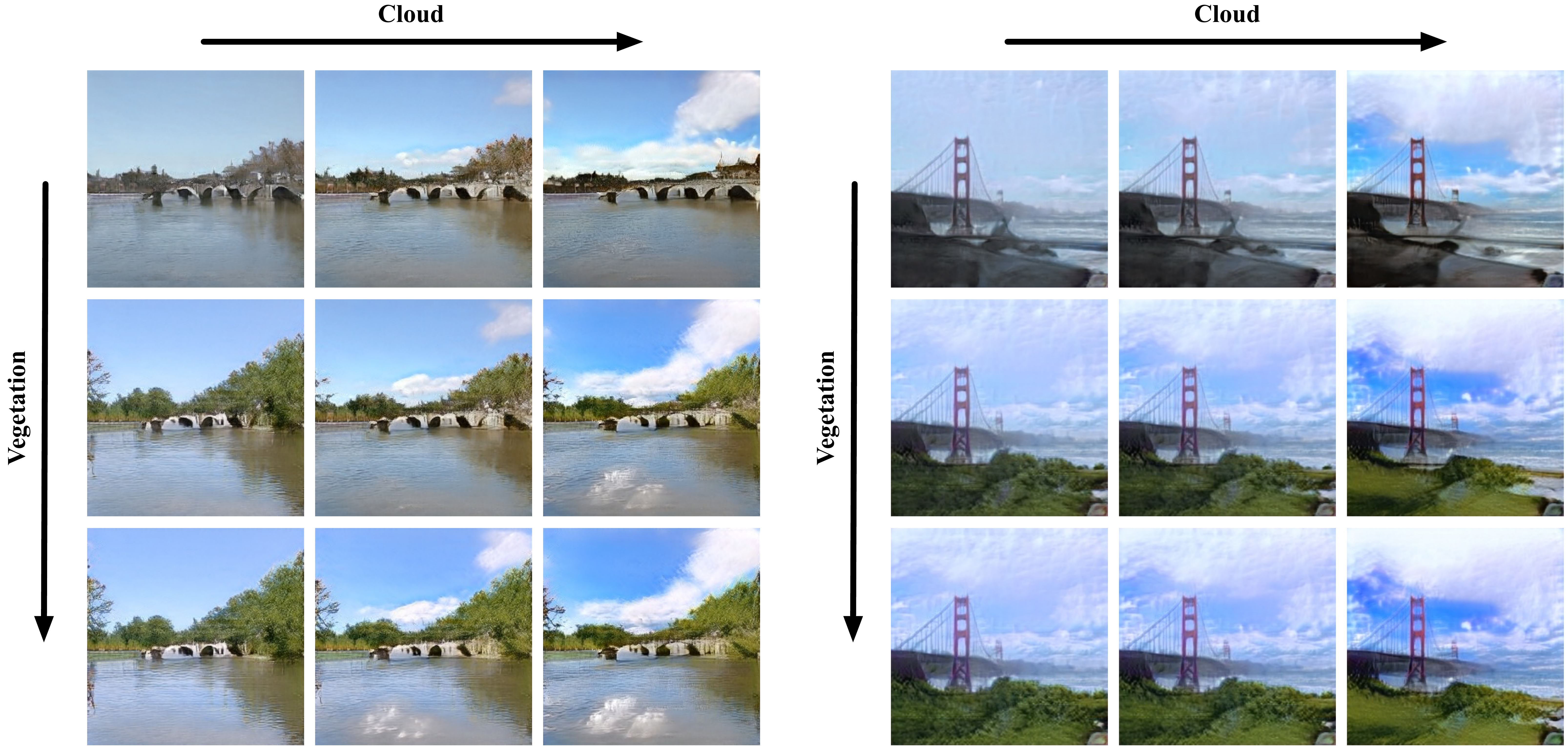}
  \captionsetup{font=small}
  \caption{
    \emph{Joint} manipulation results along both \emph{cloud} and \emph{vegetation} boundaries with bridge synthesis model.
    Along the vertical and horizontal axis, the original synthesis (the central image) is manipulated with respect to \emph{vegetation} and \emph{cloud} attributes respectively. 
  }
  \label{fig:twob}
\end{figure*}

\begin{figure*}[t]
  \centering
  \includegraphics[width=1.0\textwidth]{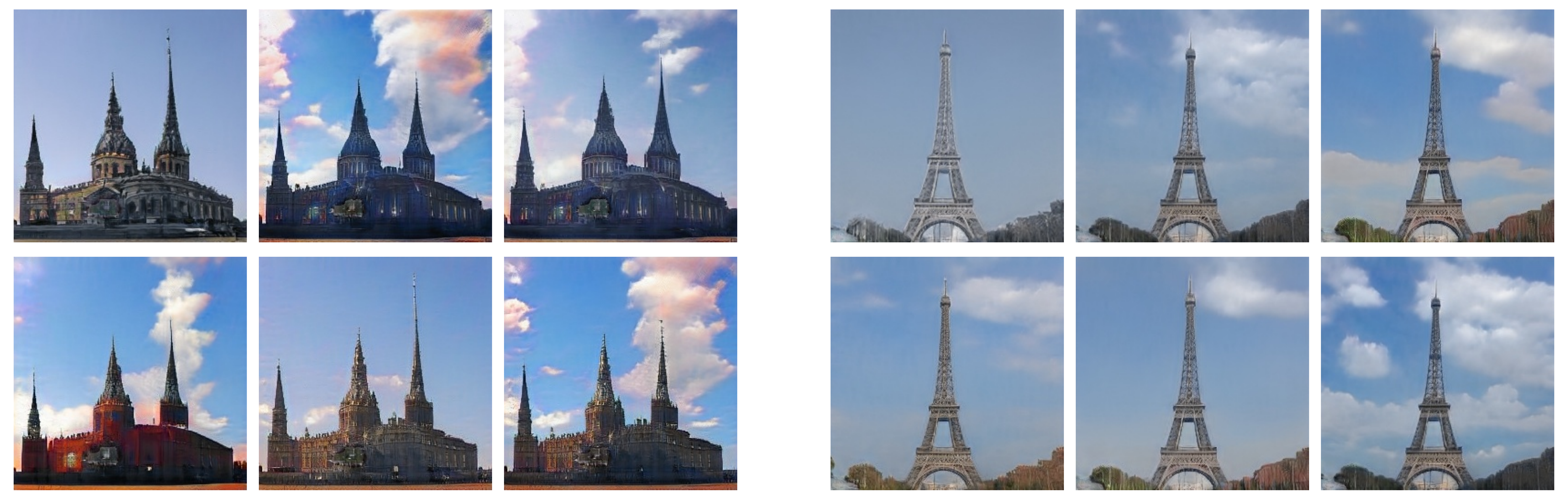}
  \captionsetup{font=small}
  \caption{
    \emph{Jittering} manipulation results with tower synthesis model for \emph{cloud} attribute. Specifically, the movement in the latent space of synthesized image is disturbed. Thus, when the cloud appears, both the shape of added cloud and appearance of the generated tower change. The top left image of two samples is the original output while the rest are the results under jittering manipulation separately. 
  }
  \label{fig:rw}
\end{figure*}

\subsection{What Makes a Scene?}\label{subsec:categorical-analysis}
As mentioned above, GAN models for synthesizing scenes are capable of encoding hierarchical semantics inside the generative representation, \emph{i.e.}, from layout, object (category), to scene attribute and color scheme.
One of the most noticeable properties is that the middle layers of GAN actually synthesize different objects for different scene categories.
It raises the question of what makes a scene as living room rather than bedroom. Thus we further dive into the encoding of categorical information in GANs, to quantify how GAN interprets a scene category as well as how the scene category is transformed from an object perspective.

We employ the StyleGAN model trained on the mixture of bedroom, living room, and dining room, and then search the semantic boundary between every two categories.
To extract the objects from the synthesized images, we apply a semantic segmentation model \cite{xiao2018unified}, which can segment 150 objects (tv, sofa, \emph{etc}) and stuff (ceiling, floor, \emph{etc}).
Specifically, we first randomly synthesize 500 living room images, and then vary the corresponding latent codes towards the ``living room-bedroom'' boundary and ``bedroom-dining room'' boundary in turn.
We segment the images before and after manipulation to get the segmentation masks, as shown in Fig.\ref{fig:seg-example}.
After tracking label mapping for each pixel via the image coordinate during the manipulation process, we are able to compute the statistics on how objects are transformed along with category changing and observe how the objects change when category is transformed.

Fig.\ref{fig:seg-example} shows the objects mapping in the category transformation process.
We can see that
(1) When an image is manipulated among different categories, most of the stuff classes (\emph{e.g.}, ceiling and floor) remain the same, but some objects are mapped into other classes.
For example, the sofa in living room is mapped to the pillow and bed in bedroom, and the bed in bedroom is further mapped to the table and chair in dining room.
This phenomenon happens because sofa, bed, dining table and chair are distinguishable objects for living room, bedroom, and dining room respectively.
Thus, when category is transformed, the representative objects are supposed to change.
%
% Thus in order to make the synthesized image look like the desired category, the model learns to render the most relevant objects.
%
(2) Some objects are sharable between different scene categories, and the GAN model is able to spot such property and learn to generate these shared objects across different classes.
For example, the lamp in living room (on the left boundary of the image) still remains after the image is converted to bedroom, especially in the same position.
%
%That is because lamp is a common object for living room and bedroom, and GAN learns to capture this information.
%
(3) With the ability to learn object mapping as well as share objects across different classes, we are able to turn an unconditional GAN into a GAN that can control category.
Typically, to make GAN produce images from different categories, class labels have to be fed into the generator to learn a categorical embedding, like BigGAN \cite{biggan}.
Our result suggests an alternative approach.

\subsection{Diverse Attribute Manipulation}\label{subsec:attribute-identification}

\noindent\textbf{Attribute Identification.}
The emergence of variation factors for scene synthesis depends on the training data.
Here we apply our method to a collection of StyleGAN models, to capture a wide range of manipulatable attributes out of the 102 scene attributes pre-defined in SUN attribute database \cite{sceneattribute}.
Each styleGAN in the collection is trained to synthesize scene images from a certain category, including both outdoor (bridge, church, tower) and indoor scenes (living room, kitchen).
Fig.\ref{fig:scene-comparison-attribute} shows the top-10 relevant semantics to each model.
We can see that ``sunny'' has high scores on all outdoor categories, while ``lighting'' has high scores on all indoor categories.
Furthermore, ``boating'' is identified for bridge model, ``touring'' for church and tower, ``reading'' for living room, ``eating'' for kitchen, and ``socializing'' for restaurant.
These results are highly consistent with human perception, suggesting the effectiveness of the proposed quantification method.

\begin{figure*}[t]
  \centering
  \includegraphics[width=1.0\textwidth]{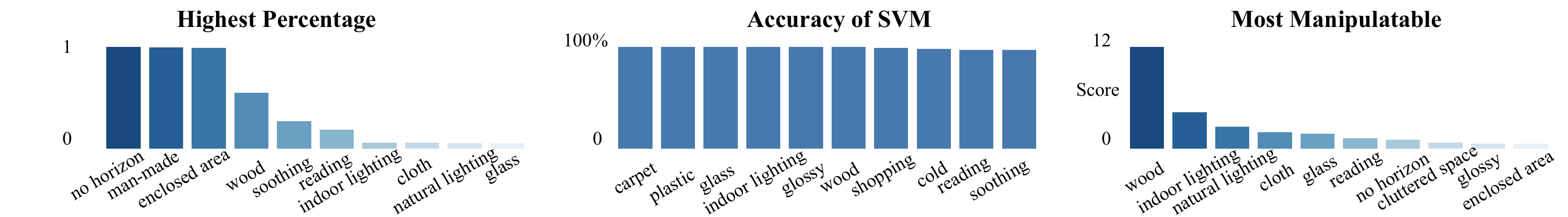}
  \captionsetup{font=small}
  \caption{
    Ablation study on the proposed re-scoring technique with StyleGAN model for bedroom synthesis.
    The left shows the percentage of scene attributes with the positive scores,
    the middle figure sorts by the accuracy of SVM classifiers, 
    while the right figure sorts by our methods.
  }
  \label{fig:ablation-study}
\end{figure*}

\begin{figure*}[t]
  \centering
  \includegraphics[width=1.0\textwidth]{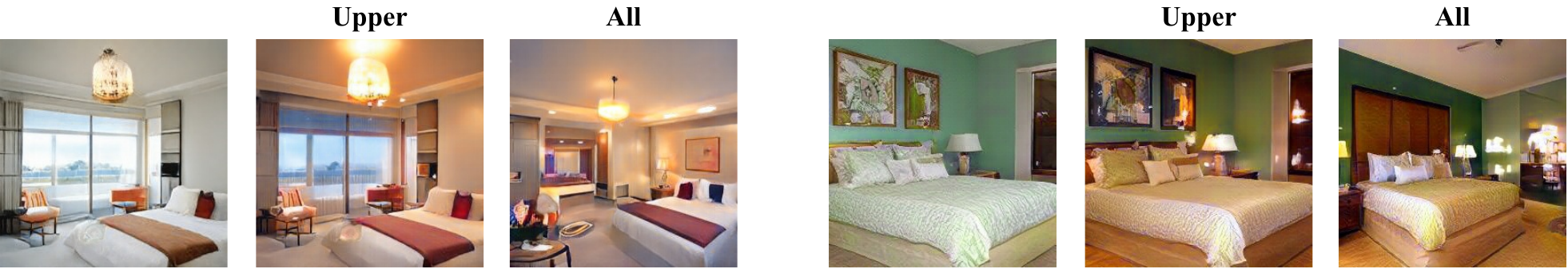}
  \captionsetup{font=small}
  \caption{
    Comparison results between manipulating latent codes at only upper (attribute-relevant) layers and manipulating latent codes at all layers with respect to \emph{indoor lighting} on StyleGAN.
  }
  \label{fig:layerwise}
\end{figure*}
\begin{figure*}[t]
  \centering
  \includegraphics[width=1.0\textwidth]{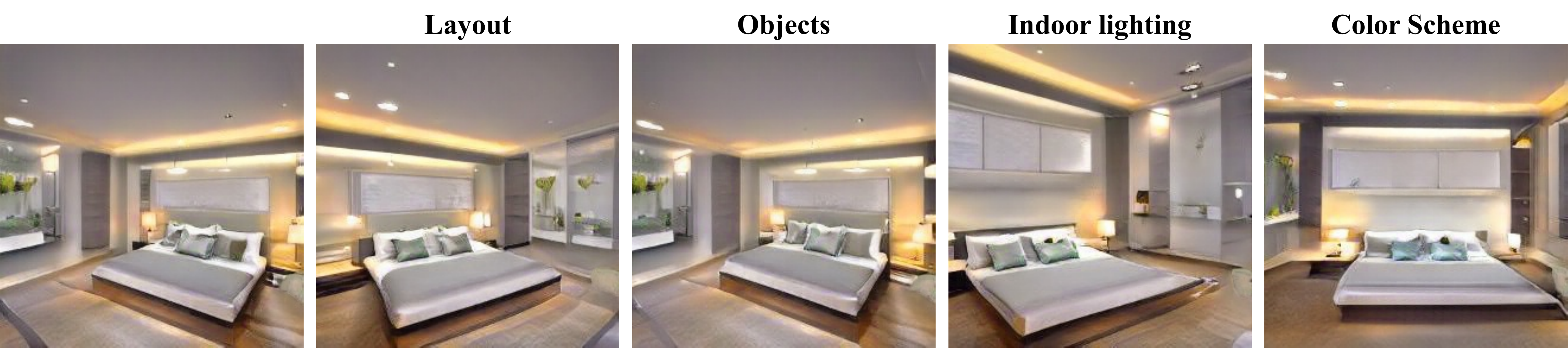}
  \captionsetup{font=small}
  \caption{
    Manipulation at the \emph{bottom} layers in 4 different directions, along the directions of \emph{layout}, \emph{objects (category)}, \emph{indoor lighting}, and \emph{color scheme} on StyleGAN.
  }
  \label{fig:directions}
\end{figure*}

\noindent\textbf{Attribute Manipulation.}
Recall the three types of manipulation in Sec.\ref{subsec:manipulating}: \emph{independent} manipulation, \emph{joint} manipulation, and \emph{jittering} manipulation.
We first conduct independent manipulation on 3 indoor and 3 outdoor scenes with the most relevant scene attributes identified with our approach.
Fig.\ref{fig:outdoor_manipulation} shows the results where the original synthesis (left image in each pair) is manipulated along the positive (right) direction.
We can tell that the edited images are still with high quality and the target attributes indeed change as desired.
We then jointly manipulate two attributes with bridge synthesis model as shown in Fig.\ref{fig:twob}.
The central image of the $3\times3$ image grid is the original synthesis, the second row and the second column show the independent manipulation results with respect to ``vegetation'' and ``cloud'' attributes respectively, while other images on the four corners are the joint manipulation results.
It turns out that we achieve good control of these two semantics and they seem to barely affect each other.
However, not all variation factors show such strong disentanglement.
From this point of view, our approach also provides a new metric to help measure the entanglement between two variation factors, which will be discused in Sec.\ref{sec:discussion}.
Finally, we evaluate the proposed \emph{jittering} manipulation by introducing noise into the ``cloud'' manipulation .
From Fig.\ref{fig:rw}, we observe that the newly introduced noise indeed increases the manipulation diversity.
It is interesting that the introduced randomness may not only affect the shape of added cloud, but also change the appearance of the synthesized tower.
But both cases keep the primary goal, which is to edit the cloudness.

\begin{figure*}[t]
  \centering
  \includegraphics[width=1.0\textwidth]{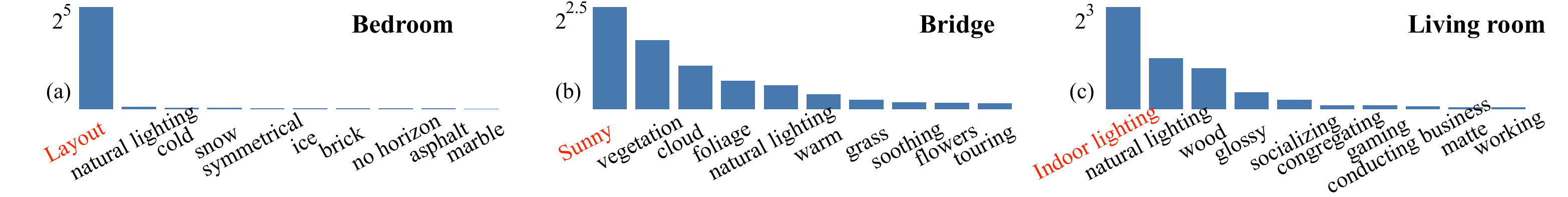}
  \captionsetup{font=small}
  \caption{
    Effects on scene attributes (already sorted) when varying a particular variation factor (in \textcolor{red}{\textbf{red}} color).
    %.
    Vertical axis shows the perturbation score $\Delta s_i$ in log scale
  }
  \label{fig:disentanglement}
\end{figure*}

\begin{figure*}[t]
  \centering
  \includegraphics[width=1.0\textwidth]{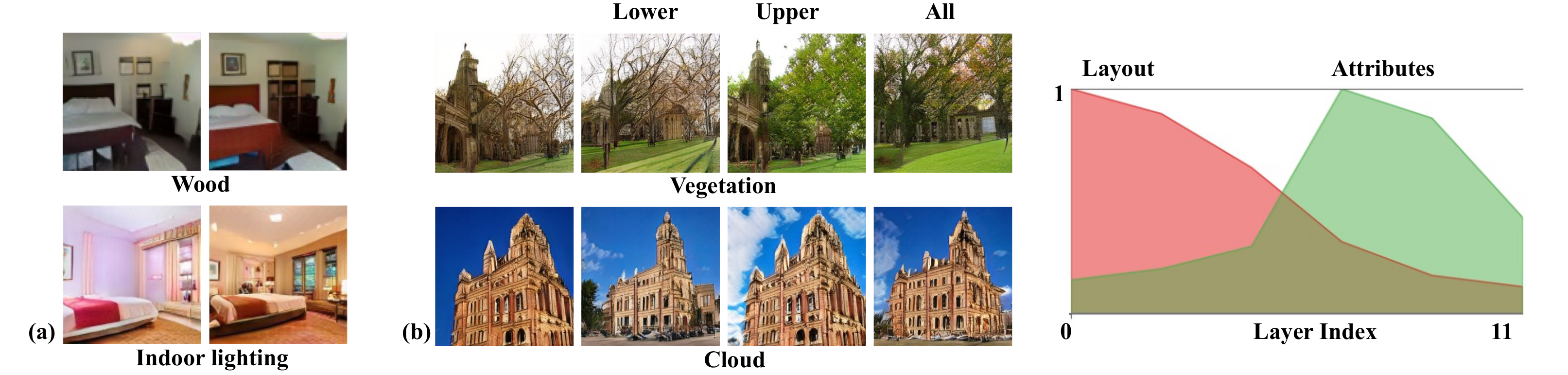}
  \captionsetup{font=small}
  \caption{
    (a) Some variation factors identified from PGGAN (bedroom).
    (b) Layer-wise analysis on BigGAN from attribute level.
  }
  \label{fig:other-models}
\end{figure*}

\subsection{Ablation Studies}\label{subsec:ablation}
\noindent\textbf{Re-scoring Technique.}
Before performing the proposed re-scoring technique, we have two more steps, which are (1) assigning semantic scores for synthesized samples, and (2) training SVM classifiers to search semantic boundary.
We would like to verify the essentiality of the re-scoring technique in identifying manipulatable semantics.
We conduct ablation study on the StyleGAN model trained for synthesizing bedrooms.
As shown in Fig.\ref{fig:ablation-study}, the left figure sorts the scene attributes by how many samples are labelled as positive ones, the middle figure sorts by the accuracy of the trained SVM classifiers, while the right figure sorts by our proposed quantification metric.

In left figure, ``no horizon'', ``man-made'', and ``enclosed area'' are attributes with highest percentage.
However, all these three attributes are default properties of the bedroom and thus not manipulatable.
On the contrary, with the re-scoring technique for verification, our method successfully filters out these invariable candidates and reveals more meaningful semantics, like ``wood'' and ``indoor lighting''.
In addition, our method also manages to identify some less frequent but actually manipulatable scene attributes, such as ``cluttered space''.

In the middle figure, almost all attributes get similar scores, making them indistinguishable.
Actually, even the worst SVM classifier (\emph{i.e.}, ``railroad'') achieves 72.3\% accuracy.
That is because even some variation factors are not encoded in the latent representation (or say, not manipulatable), the corresponding attribute classifier still assigns synthesized images with different scores.
Training SVM on these inaccurate data can also result in a separation boundary, even it is not expected as the target concept.
Therefore, only relying on the SVM classifier is not enough to detect relevant variation factors.
By contrast, our method pays more attention to the score modulation after varying the latent code, which is not biased by the initial response of attribute classifier or the performance of SVM.
As a result, we are able to thoroughly yet precisely detect the variation factors in the latent space from a broad candidate set.

\noindent\textbf{Layer-wise Manipulation.}
To further validate the emergence of semantic hierarchy, we make ablation study on layer-wise manipulation with StyleGAN model.
First, we select ``indoor lighting'' as the target semantic, and vary the latent code only on upper (attribute-relevant) layers \emph{v.s.} on all layers.
We can easily tell from Fig.\ref{fig:layerwise} that when manipulation ``indoor lighting'' at all layers, the objects inside the room are also changed.
By contrast, manipulating latent codes only at attribute-relevant layers can satisfyingly increase the indoor lighting without affecting other factors.
Second, we select bottom layers as the target layers, and select boundaries from all four abstraction levels for manipulation.
As shown in Fig.\ref{fig:directions}, no matter what level of semantics we choose, as long as the latent code is modified at bottom (layout-relevant) layers, only layout instead of all other semantics varies.
These two experiments further verify our discovery about the emergence of the semantic hierarchy that the early layers tend to determine the spatial layout and configuration instead of other abstraction level semantics.

% !TEX root = main.tex

\section{Discussions}\label{sec:discussion}

\noindent\textbf{Disentanglement of Semantics.}
Some variation factors we detect in the generative representation are more disentangled with each other than other semantics.
Compared to the perceptual path length and linear separability described in \citet{stylegan} and the cosine similarity proposed in \citet{shen2019interpreting}, our work offers a new metric for disentanglement analysis.
In particular, we move the latent code along one semantic direction and then check how the semantic scores of other factors change accordingly.
As shown in Fig.\ref{fig:disentanglement}(a), when we modify the spatial layout, all scene attributes are barely affected, suggesting that GAN learns to disentangle layout-level semantic from attribute-level.
However, there are also some scene attributes (from same abstraction level) entangling with each other.
Taking Fig.\ref{fig:disentanglement}(c) as an example, when modulating ``indoor lighting'', ``natural lighting'' also varies.
This is also aligned with human perception, further demonstrating the effectiveness of our proposed quantification metric.

\noindent\textbf{Application to Other GANs.}
We further apply our method for two other GAN structures, \emph{i.e.}, PGGAN \cite{pggan} and BigGAN \cite{biggan}.
These two models are trained on LSUN dataset \citet{yu2015lsun} and Places dataset \cite{zhou2017places} respectively.
Compared to StyleGAN, PGGAN feeds the latent vector only to the very first convolutional layer and hence does not support layer-wise analysis.
But the proposed re-scoring method can still be applied to help identify manipulatable semantics, as shown in Fig.\ref{fig:other-models}(a).
BigGAN is the state-of-the-art conditional GAN model that concatenates the latent vector with a class-guided embedding code before feeding it to the generator, and it also allows layer-wise analysis like StyleGAN.
Fig.\ref{fig:other-models}(b) gives analysis results on BigGAN from attribute level,
%\footnote{We do not analyze BigGAN from layout and category level, since BigGAN employs an additional vector to encode category information and is trained on both indoor and outdoor scenes where layout is not well-defined.}.
%
where we can tell that scene attribute can be best modified at upper layers compared to lower layers or all layers.
As for BigGAN model with $256\times256$ resolution, there are total 12 convolutional layers.
As the category information is already encoded in the ``class'' code, we only separate the layers to two groups, which are \emph{lower} (bottom 6 layers) and \emph{upper} (top 6 layers).
Meanwhile, the quantitative curve shows the consistent result with the discovery on StyleGAN as in Fig.\ref{fig:layer-wise-analysis}(a).
These results demonstrate the generalization ability of our approach as well as the emergence of manipulatable factors in other GANs.

\noindent\textbf{Limitation.}
%
%Despite the success of our proposed re-scoring technique in quantitatively identifying the hierarchical manipulatable latent variation factors in the deep generative representations,
There are several limitations for future improvement.
First, the layout classifier can only detect the layout structure of indoor scenes.
But for a more general analysis on both indoor and outdoor scene categories, there lacks a unified definition of the spatial layout.
For example, our framework cannot change the layout of outdoor church images. In future work, we will leverage the computational photography tools that recover the 3D camera pose of the image, thus we can extract more universal viewpoint representation for the synthesized images.
Second, our proposed re-scoring technique relies on the performances of the off-the-shelf classifiers.
For some of the attributes, the classifiers are not so accurate, which leads to poor manipulation boundary.
This problem could be addressed with more powerful discriminative models.
Third, for simplicity we only use the linear SVM for semantic boundary search.
This limits our framework from interpreting the latent semantic subspace with more complex and nonlinear structure.

\section{Conclusion}
In this paper, we show the emergence of highly-structured variation factors inside the deep generative representations learned by GANs with layer-wise stochasticity.
In particular, the GAN model spontaneously learns to set up layout at early layers, generate categorical objects at middle layers, and render scene attribute and color scheme at later layers when trained to synthesize scenes.
A re-scoring method is proposed to quantitatively identify the manipulatable semantic concepts within a well-trained model, enabling photo-realistic scene manipulation. We will explore to extend this manipulation capability of GANs for real image editing in future work. 

\noindent \textbf{Acknowledgement:} This work is supported by the Early Career Scheme of Hong Kong (No. 24206219) and RSFS grant from CUHK Faculty of Engineering (No. 3133233).

\bibliographystyle{spbasic}      % basic style, author-year citations
\bibliography{references}

\begin{thebibliography}{43}
\providecommand{\natexlab}[1]{#1}
\providecommand{\url}[1]{{#1}}
\providecommand{\urlprefix}{URL }
\expandafter\ifx\csname urlstyle\endcsname\relax
  \providecommand{\doi}[1]{DOI~\discretionary{}{}{}#1}\else
  \providecommand{\doi}{DOI~\discretionary{}{}{}\begingroup
  \urlstyle{rm}\Url}\fi
\providecommand{\eprint}[2][]{\url{#2}}

\bibitem[{Agrawal et~al(2014)Agrawal, Girshick, and
  Malik}]{agrawal2014analyzing}
Agrawal P, Girshick R, Malik J (2014) Analyzing the performance of multilayer
  neural networks for object recognition. In: ECCV

\bibitem[{Alain and Bengio(2016)}]{alain2016understanding}
Alain G, Bengio Y (2016) Understanding intermediate layers using linear
  classifier probes. arXiv:161001644

\bibitem[{Bau et~al(2017)Bau, Zhou, Khosla, Oliva, and
  Torralba}]{bau2017network}
Bau D, Zhou B, Khosla A, Oliva A, Torralba A (2017) Network dissection:
  Quantifying interpretability of deep visual representations. In: CVPR

\bibitem[{Bau et~al(2019)Bau, Zhu, Strobelt, Zhou, Tenenbaum, Freeman, and
  Torralba}]{gandissection}
Bau D, Zhu JY, Strobelt H, Zhou B, Tenenbaum JB, Freeman WT, Torralba A (2019)
  Gan dissection: Visualizing and understanding generative adversarial
  networks. In: ICLR

\bibitem[{Bengio et~al(2013)Bengio, Courville, and
  Vincent}]{bengio2013representation}
Bengio Y, Courville A, Vincent P (2013) Representation learning: A review and
  new perspectives. TPAMI

\bibitem[{Brock et~al(2019)Brock, Donahue, and Simonyan}]{biggan}
Brock A, Donahue J, Simonyan K (2019) Large scale gan training for high
  fidelity natural image synthesis. In: ICLR

\bibitem[{Cheng et~al(2014)Cheng, Zheng, Lin, Vineet, Sturgess, Crook, Mitra,
  and Torr}]{cheng2014imagespirit}
Cheng MM, Zheng S, Lin WY, Vineet V, Sturgess P, Crook N, Mitra NJ, Torr P
  (2014) Imagespirit: Verbal guided image parsing. ACM Trans on Graphics

\bibitem[{Choi et~al(2018)Choi, Choi, Kim, Ha, Kim, and Choo}]{choi2018stargan}
Choi Y, Choi M, Kim M, Ha JW, Kim S, Choo J (2018) Stargan: Unified generative
  adversarial networks for multi-domain image-to-image translation. In: CVPR

\bibitem[{Goetschalckx et~al(2019)Goetschalckx, Andonian, Oliva, and
  Isola}]{ganalyze}
Goetschalckx L, Andonian A, Oliva A, Isola P (2019) Ganalyze: Toward visual
  definitions of cognitive image properties. In: ICCV

\bibitem[{Gonzalez-Garcia et~al(2018)Gonzalez-Garcia, Modolo, and
  Ferrari}]{gonzalez2018semantic}
Gonzalez-Garcia A, Modolo D, Ferrari V (2018) Do semantic parts emerge in
  convolutional neural networks? IJCV

\bibitem[{Goodfellow et~al(2014)Goodfellow, Pouget-Abadie, Mirza, Xu,
  Warde-Farley, Ozair, Courville, and Bengio}]{gan}
Goodfellow I, Pouget-Abadie J, Mirza M, Xu B, Warde-Farley D, Ozair S,
  Courville A, Bengio Y (2014) Generative adversarial nets. In: NeurIPS

\bibitem[{Heusel et~al(2017)Heusel, Ramsauer, Unterthiner, Nessler, and
  Hochreiter}]{fid}
Heusel M, Ramsauer H, Unterthiner T, Nessler B, Hochreiter S (2017) Gans
  trained by a two time-scale update rule converge to a local nash equilibrium.
  In: NeurIPS

\bibitem[{Isola et~al(2017)Isola, Zhu, Zhou, and Efros}]{isola2017image}
Isola P, Zhu JY, Zhou T, Efros AA (2017) Image-to-image translation with
  conditional adversarial networks. In: CVPR

\bibitem[{Jahanian et~al(2020)Jahanian, Chai, and Isola}]{steerability}
Jahanian A, Chai L, Isola P (2020) On the''steerability" of generative
  adversarial networks. ICLR

\bibitem[{Karacan et~al(2016)Karacan, Akata, Erdem, and
  Erdem}]{karacan2016learning}
Karacan L, Akata Z, Erdem A, Erdem E (2016) Learning to generate images of
  outdoor scenes from attributes and semantic layouts. arXiv preprint
  arXiv:161200215

\bibitem[{Karras et~al(2018)Karras, Aila, Laine, and Lehtinen}]{pggan}
Karras T, Aila T, Laine S, Lehtinen J (2018) Progressive growing of gans for
  improved quality, stability, and variation. In: ICLR

\bibitem[{Karras et~al(2019)Karras, Laine, and Aila}]{stylegan}
Karras T, Laine S, Aila T (2019) A style-based generator architecture for
  generative adversarial networks. In: CVPR

\bibitem[{Laffont et~al(2014)Laffont, Ren, Tao, Qian, and
  Hays}]{laffont2014transient}
Laffont PY, Ren Z, Tao X, Qian C, Hays J (2014) Transient attributes for
  high-level understanding and editing of outdoor scenes. ACM Trans on Graphics

\bibitem[{Liao et~al(2017)Liao, Yao, Yuan, Hua, and Kang}]{liao2017visual}
Liao J, Yao Y, Yuan L, Hua G, Kang SB (2017) Visual attribute transfer through
  deep image analogy. arXiv preprint arXiv:170501088

\bibitem[{Luan et~al(2017)Luan, Paris, Shechtman, and Bala}]{luan2017deep}
Luan F, Paris S, Shechtman E, Bala K (2017) Deep photo style transfer. In: CVPR

\bibitem[{Mahendran and Vedaldi(2015)}]{mahendran2015understanding}
Mahendran A, Vedaldi A (2015) Understanding deep image representations by
  inverting them. In: CVPR

\bibitem[{Morcos et~al(2018)Morcos, Barrett, Rabinowitz, and
  Botvinick}]{morcos2018importance}
Morcos AS, Barrett DG, Rabinowitz NC, Botvinick M (2018) On the importance of
  single directions for generalization. In: ICLR

\bibitem[{Nguyen et~al(2016)Nguyen, Dosovitskiy, Yosinski, Brox, and
  Clune}]{nguyen2016synthesizing}
Nguyen A, Dosovitskiy A, Yosinski J, Brox T, Clune J (2016) Synthesizing the
  preferred inputs for neurons in neural networks via deep generator networks.
  In: NeurIPS

\bibitem[{Nguyen-Phuoc et~al(2019)Nguyen-Phuoc, Li, Theis, Richardt, and
  Yang}]{nguyen2019hologan}
Nguyen-Phuoc T, Li C, Theis L, Richardt C, Yang YL (2019) Hologan: Unsupervised
  learning of 3d representations from natural images. In: ICCV

\bibitem[{Park et~al(2019)Park, Liu, Wang, and Zhu}]{park2019semantic}
Park T, Liu MY, Wang TC, Zhu JY (2019) Semantic image synthesis with
  spatially-adaptive normalization. In: CVPR

\bibitem[{Patterson et~al(2014)Patterson, Xu, Su, and Hays}]{sceneattribute}
Patterson G, Xu C, Su H, Hays J (2014) The sun attribute database: Beyond
  categories for deeper scene understanding. IJCV

\bibitem[{Radford et~al(2016)Radford, Metz, and Chintala}]{dcgan}
Radford A, Metz L, Chintala S (2016) Unsupervised representation learning with
  deep convolutional generative adversarial networks. In: ICLR

\bibitem[{Shaham et~al(2019)Shaham, Dekel, and Michaeli}]{shaham2019singan}
Shaham TR, Dekel T, Michaeli T (2019) Singan: Learning a generative model from
  a single natural image. In: ICCV

\bibitem[{Shen et~al(2018)Shen, Luo, Yan, Wang, and Tang}]{shen2018faceid}
Shen Y, Luo P, Yan J, Wang X, Tang X (2018) Faceid-gan: Learning a symmetry
  three-player gan for identity-preserving face synthesis. In: CVPR

\bibitem[{Shen et~al(2019)Shen, Gu, Tang, and Zhou}]{shen2019interpreting}
Shen Y, Gu J, Tang X, Zhou B (2019) Interpreting the latent space of gans for
  semantic face editing. arXiv preprint arXiv:190710786

\bibitem[{Simonyan et~al(2014)Simonyan, Vedaldi, and
  Zisserman}]{simonyan2013deep}
Simonyan K, Vedaldi A, Zisserman A (2014) Deep inside convolutional networks:
  Visualising image classification models and saliency maps. In: ICLR Workshop

\bibitem[{Wang et~al(2018)Wang, Liu, Zhu, Tao, Kautz, and
  Catanzaro}]{wang2018high}
Wang TC, Liu MY, Zhu JY, Tao A, Kautz J, Catanzaro B (2018) High-resolution
  image synthesis and semantic manipulation with conditional gans. In: CVPR

\bibitem[{Xiao et~al(2010)Xiao, Hays, Ehinger, Oliva, and
  Torralba}]{sundatabase}
Xiao J, Hays J, Ehinger KA, Oliva A, Torralba A (2010) Sun database:
  Large-scale scene recognition from abbey to zoo. In: CVPR

\bibitem[{Xiao et~al(2018{\natexlab{a}})Xiao, Hong, and Ma}]{xiao2018elegant}
Xiao T, Hong J, Ma J (2018{\natexlab{a}}) Elegant: Exchanging latent encodings
  with gan for transferring multiple face attributes. In: ECCV

\bibitem[{Xiao et~al(2018{\natexlab{b}})Xiao, Liu, Zhou, Jiang, and
  Sun}]{xiao2018unified}
Xiao T, Liu Y, Zhou B, Jiang Y, Sun J (2018{\natexlab{b}}) Unified perceptual
  parsing for scene understanding. In: ECCV

\bibitem[{Yao et~al(2018)Yao, Hsu, Zhu, Wu, Torralba, Freeman, and
  Tenenbaum}]{yao20183d}
Yao S, Hsu TM, Zhu JY, Wu J, Torralba A, Freeman B, Tenenbaum J (2018) 3d-aware
  scene manipulation via inverse graphics. In: NeurIPS

\bibitem[{Yosinski et~al(2014)Yosinski, Clune, Bengio, and
  Lipson}]{yosinski2014transferable}
Yosinski J, Clune J, Bengio Y, Lipson H (2014) How transferable are features in
  deep neural networks? In: NeurIPS

\bibitem[{Yu et~al(2015)Yu, Seff, Zhang, Song, Funkhouser, and
  Xiao}]{yu2015lsun}
Yu F, Seff A, Zhang Y, Song S, Funkhouser T, Xiao J (2015) Lsun: Construction
  of a large-scale image dataset using deep learning with humans in the loop.
  arXiv preprint arXiv:150603365

\bibitem[{Zeiler and Fergus(2014)}]{zeiler2014visualizing}
Zeiler MD, Fergus R (2014) Visualizing and understanding convolutional
  networks. In: ECCV

\bibitem[{Zhang et~al(2019)Zhang, Zhang, and Gu}]{layout}
Zhang W, Zhang W, Gu J (2019) Edge-semantic learning strategy for layout
  estimation in indoor environment. In: IEEE Transactions on Cybernetics

\bibitem[{Zhou et~al(2015)Zhou, Khosla, Lapedriza, Oliva, and
  Torralba}]{zhou2014object}
Zhou B, Khosla A, Lapedriza A, Oliva A, Torralba A (2015) Object detectors
  emerge in deep scene cnns. In: ICLR

\bibitem[{Zhou et~al(2017)Zhou, Lapedriza, Khosla, Oliva, and
  Torralba}]{zhou2017places}
Zhou B, Lapedriza A, Khosla A, Oliva A, Torralba A (2017) Places: A 10 million
  image database for scene recognition. TPAMI

\bibitem[{Zhu et~al(2017)Zhu, Park, Isola, and Efros}]{zhu2017unpaired}
Zhu JY, Park T, Isola P, Efros AA (2017) Unpaired image-to-image translation
  using cycle-consistent adversarial networks. In: ICCV

\end{thebibliography}


\end{document}